\documentclass{article}

\usepackage{microtype}
\usepackage{graphicx}
\usepackage{subfigure}
\usepackage{booktabs} % for professional tables

\usepackage{hyperref}

\usepackage[accepted]{icml2023}

\usepackage{amsmath}
\usepackage{amssymb}
\usepackage{mathtools}
\usepackage{amsthm}

\theoremstyle{plain}
\newtheorem{theorem}{Theorem}[section]
\newtheorem{proposition}[theorem]{Proposition}

\newtheorem{corollary}[theorem]{Corollary}
\theoremstyle{definition}
\newtheorem{definition}[theorem]{Definition}

\theoremstyle{remark}

\usepackage[utf8]{inputenc}
\usepackage[T1]{fontenc}    % use 8-bit T1 fonts
\usepackage{dsfont} % bold digits such as indicator function
\usepackage{xcolor}
\usepackage{enumitem} % -- TV -- Custom commands
\usepackage{wrapfig}
\usepackage{bm} % bold Greek math symbols
\usepackage{multirow}
\usepackage{fancybox}

\usepackage{pgfplots}
\usepackage{pgfplotstable}
\usepgfplotslibrary{groupplots} % for several plot on same figure
\usepgfplotslibrary{fillbetween}
\pgfplotsset{compat=1.16}
\pgfplotsset{grid = major, grid style={gray!30!white}}

\definecolor{darkgreen}{rgb}{0.31, 0.47, 0.26}

\DeclareMathOperator*{\argmin}{argmin}

\DeclareMathOperator{\TruncNorm}{TruncNormal}
\DeclareMathOperator{\Exp}{\mathbb{E}}
\newcommand{\xAlt}{x^{\text{alt}}}
\newcommand{\zAlt}{z^{\text{alt}}}

\DeclareMathOperator{\CVaR}{CVaR_\alpha}

\icmltitlerunning{Explainable Data-Driven Optimization: From Context to Decision and Back Again}

\begin{document}

\twocolumn[
\icmltitle{Explainable Data-Driven Optimization:\\From Context to Decision and Back Again}

\begin{icmlauthorlist}
    \icmlauthor{Alexandre Forel}{polymagi}
    \icmlauthor{Axel Parmentier}{cermics}
    \icmlauthor{Thibaut Vidal}{polymagi}
\end{icmlauthorlist}

\icmlaffiliation{polymagi}{CIRRELT \& SCALE-AI Chair in Data-Driven Supply Chains, Department of Mathematical and Industrial Engineering, Polytechnique Montreal, Montreal, Canada}
\icmlaffiliation{cermics}{CERMICS, \'Ecole des Ponts, Marne-la-Vall\'ee, France}

\icmlcorrespondingauthor{Alexandre Forel}{alexandre.forel@polymtl.ca}

\icmlkeywords{Data-driven optimization; contextual stochastic problems; interpretability}

\vskip 0.3in
]

\printAffiliationsAndNotice{}  % leave blank if no need to mention equal contribution

\begin{abstract}
    Data-driven optimization uses contextual information and machine learning algorithms to find solutions to decision problems with uncertain parameters. While a vast body of work is dedicated to interpreting machine learning models in the classification setting, explaining decision pipelines involving learning algorithms remains unaddressed. This lack of interpretability can block the adoption of data-driven solutions as practitioners may not understand or trust the recommended decisions. We bridge this gap by introducing a counterfactual explanation methodology tailored to explain solutions to data-driven problems. We introduce two classes of explanations and develop methods to find nearest explanations of random forest and nearest-neighbor predictors. We demonstrate our approach by explaining key problems in operations management such as inventory management and routing.
\end{abstract}
\section{Introduction}
\label{sec:intro}
Data-driven optimization leverages contextual information to solve problems subject to uncertainty by combining machine learning and optimization methods. Contextual information includes auxiliary data such as prices, temporal information, or meteorological data. While utilizing contextual information can significantly improve data-driven decision-making \citep{Hannah2010, Bertsimas2020, Misic2020}, the resulting decision pipelines are often complex and lack transparency. Yet, decisions must be interpretable to be used in practice. They should allow practitioners to understand what features of the context make a specific decision optimal, and to what extent a change in the context leads to changes in the decisions. This is especially relevant in industrial settings when comparing a new data-driven policy to the existing policy, which might be based on a mix of expert knowledge and quantitative methods.

To enable explainable data-driven optimization, we revisit the concept of counterfactual explanation, used extensively to explain machine learning classifiers \citep{Wachter2017, Verma2020}. Thus, we explain a decision by answering the question: \textbf{In what alternative context would the previous expert-based solution be better than the data-driven solution?} This alternative context forms a contrastive explanation, highlighting the main features that make the data-driven decision optimal. In other words, we focus on explaining decisions rather than classes or labels, as is typical in explainable AI. A notable difference is that decision spaces often involve an intractable number of possible decisions (rather than a few classes), making the explanation task much harder.

\begin{figure}[hb]
    \centering
    \includegraphics[trim=2mm 2mm 2mm 2mm, clip, width=0.98\linewidth]{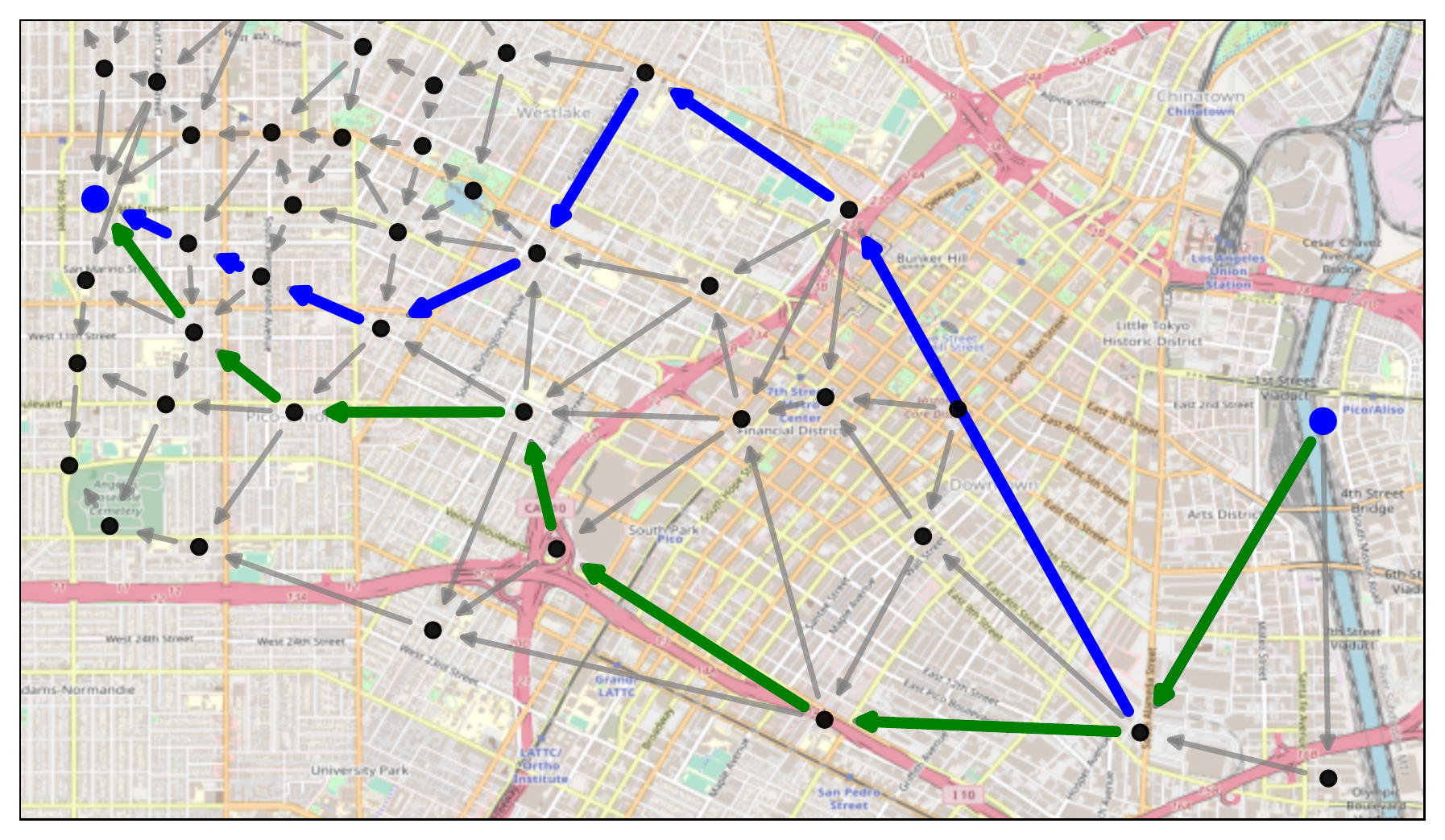}
    \caption{Shortest path over Los Angeles downtown area: optimal path $z^*$ in context $x_{n+1} = \{\text{Midday}, 57.17, 4, 0, 6.99, 2, 11\}$ shown in \textbf{\textcolor{blue}{blue}}, and alternative path $\zAlt$ shown in \textbf{\textcolor{darkgreen}{green}}. The alternative path becomes optimal when a single feature of the context is modified: "Midday" $\rightarrow$ "AM". The details are given in Section~\ref{sec:uberPath}.}
    \label{fig:uberPath}
\end{figure}
To illustrate, consider a shortest-path problem over the Los Angeles downtown area. Given historical data and contextual information, a data-driven optimization model can provide a shortest path between two nodes, as shown in blue in Figure~\ref{fig:uberPath}. Typically, the number of possible shortest paths between two points grows exponentially with the size of the network. Now, suppose that a delivery company is using an alternative routing decision that takes instead the south route shown in green. We explain why the new data-driven path is optimal by providing a counterfactual explanation: the alternative company route would be better if the time of the day were early morning rather than midday. In fact, in
this alternative context, the company route is optimal.

\textbf{Background: from context to decision.} In data-driven optimization, the decision-maker minimizes a cost function $c(z, y)$ with uncertain parameters $y \in \mathcal{Y} \subset \mathbb{R}^{d_y}$ by taking a decision $z \in \mathcal{Z} \subset \mathbb{R}^{d_z}$. Contextual information~$x \in \mathcal{X} \subset \mathbb{R}^{d_x}$ is available as a vector of $d_x$ relevant features. The context and uncertain parameters are assumed to follow a joint probability distribution $(x, y) \sim P_{x,y}$ so that the optimal decision $z^*$ depends on the context $x$. The decision-maker is thus looking for the decision policy $\pi : \mathcal{X} \mapsto \mathcal{Z} $ that minimizes the expected costs $\Exp_{x,y} \left[ c\left(\pi(x), y \right) \right]$.

Given historical data $\{(x^i, y^i)\}_{i=1}^n$, machine learning predictors can be trained to infer the relationship between the context and the uncertain parameters. In particular, the predictors presented by \citet{Bertsimas2020} assign a weight $w_i \in [0,1]$ to historical observations, reflecting their similarity with the new context in which the decision is made. For any new context $x_{n+1}$, a weighted sample-average approximation (SAA) of the contextual stochastic optimization problem given by:
\begin{equation}
    \label{eq:wSAA}
    z^* = \argmin_{z \in \mathcal{Z}} \sum_{i=1}^n w_i(x_{n+1}) c(z, y^i)
\end{equation}
can be solved to obtain the contextually optimal decision~$z^*$. Random forests (RF) and nearest-neighbors (k-NN) have been shown to generalize well to unseen data \citep{Lin2022}, providing the lowest out-of-sample cost among the predictors studied by \citet{Bertsimas2020}. Data-driven decision pipelines are thus made of two layers shown in~\eqref{eq:pipeline}: a trained machine learning algorithm that returns adaptive weight functions, and an optimization layer that solves Problem~\eqref{eq:wSAA} to prescribe a decision.
\begin{equation}
    \label{eq:pipeline}
    \xrightarrow[]{\text{$x$}}
    \fbox{
        \begin{Bcenter}
            Predictor\\
            \textit{(k-NN or RF)}
        \end{Bcenter}}
    \xrightarrow[]{\text{$\{w_i(x)\}_{i=1}^n$}}
    \fbox{
        \begin{Bcenter}
            Optimization\\
            \textit{(Weighted SAA)}
        \end{Bcenter}}
    \xrightarrow[]{\text{$z^*$}}
\end{equation}

Many types of pipelines combining machine learning and optimization have been proposed recently (see, e.g., the tutorial of \citealt{chou_integrating_2022}). We focus on pipelines based on RF and k-NN because these predictors have received significant attention and are shown to be effective in the vast majority of papers on the topic. Still, our framework for explaining data-driven decisions is applicable to any integrated pipeline based on weighted SAA.

\textbf{Interpretability: back to the context.} To explain a decision, we need to reverse the data-driven pipeline in (2) and link back any alternative decision to the closest context in which it performs well. To achieve this, we make the following contributions:
\begin{enumerate}[topsep=-0.5\parskip, itemsep=3pt, wide=5pt]
    \item We extend the concept of counterfactual explanations to interpret the decisions of data-driven optimization and foster their practical adoption. We define two classes of explanations: relative and absolute, for which a given alternative decision is respectively better than the data-driven decision or optimal. We consider risk-neutral and risk-averse single and two-stage stochastic problems, and identify structural properties of the two classes of explanations.
    \item We develop algorithms to find relative and absolute explanations for random forest and nearest-neighbor predictors. Using integer-programming methods, we obtain optimal explanations in the sense that their distance to the initial context is guaranteed to be minimal. Our explanation framework is flexible and could be implemented in conjunction with any counterfactual explanation method developed in the classification setting.
    \item We demonstrate the applicability of our approach on a selection of key problems in operations management adapted from the recent literature. We show that our methods can find nearest explanations in reasonable times with up to $500$ features and $5000$ data points. In particular, relative explanations of random forest predictors can be obtained in a few seconds, even when the number of features ($d_x$), size of the training set ($n$), or dimensions of the underlying decision problem ($d_y$ and $d_z)$ are large.
\end{enumerate}
\subsection{Related Works}
\label{sec:litReview}
\textbf{Data-driven optimization.}
The core focus of data-driven decision-making is to develop sample-efficient methods for obtaining the adaptive weights $w_i(x)$. Indeed, the amount of data available for contextual stochastic optimization is often small, consisting of a few dozen to hundreds of observations. This typically depends on the decision problem at hand. For instance, retail operations may be based on weekly sales observations collected over a few years. Recent approaches use machine learning algorithms and provide theoretical guarantees on out-of-sample costs \citep{Ban2019, Bertsimas2020, Notz2022}. One notable advantage of simple models such as random forests is that they can be trained to optimize the cost of the decision problem rather than simply fitting the data at hand \citep{Kallus2022}. \citet{Lin2022} also study the value of risk aversion to avoid overfitting.

\textbf{Counterfactual explanations of classifiers.}
Since the seminal work of \citet{Wachter2017}, counterfactual explanations have received a great amount of attention and are used widely to interpret black-box classifiers \citep{Verma2020, Karimi2022}. Integer programming has proved especially valuable to explain the predictions of random forests, allowing the inclusion of complex actionability, plausibility, and robustness constraints \citep{Cui2015, Kanamori2020, Parmentier2021, Forel2022}. Interestingly, when a decision policy is given directly by a random forest \citep{Biggs2022} or a decision tree \citep{Elmachtoub2020}, traditional counterfactual explanation methods can be used straightforwardly to explain the decision policy. However, this is not the case in our setting, since the machine learning algorithm only provides weights that are used subsequently in an optimization model.

\textbf{Interpretable decision-making.} Despite high practical relevance, the interpretability of decisions obtained by solving optimization models has not received a large amount of attention. \citet{Korikov2021} study the interpretability of deterministic optimization problems through the lens of inverse optimization. Regretting the lack of interpretability of existing predictors, \citet{Notz2020} proposes an interpretable prescriptive method based on boosting, which performs slightly worse than random forest predictors. In this work, we show that it is possible to solve the interpretability problem of machine learning predictors such as random forests while retaining their state-of-the-art performance.
\section{Problem Statement and Properties}
\label{sec:problem}
Let $\{(x^i, y^i)\}_{i=1}^n$ be the set of historical observations and $\{w_i(x)\}_{i=1}^n$ be the weight functions given by a trained predictor. In this section, we introduce the random forest and nearest-neighbor predictors, present risk-neutral and risk-averse contextual stochastic problems, and formalize the problem of finding counterfactual explanations of decisions.

\subsection{Predictors and Adaptive Weights}
\label{sec:weightsDef}
A random forest predictor is made of $T$ regression trees. It can be trained following the standard procedure of \citet{Breiman2001} that minimizes the estimation error or following the approach of \citet{Kallus2022} that takes into account the cost of decisions. In both cases, the random forest weights assigned to a past observation are calculated as an average over the weights of the individual trees. The weights are given by $w_i^{RF}(x) = \frac{1}{T} \sum_{t=1}^T w_{i, t}^{DT}(x)$ where $w_{i, t}^{DT}(x)$ is the weight given to sample $i$ by the $t$-th tree in the forest. A tree assigns equal weights to all samples in the same leaf node as $x$, so that the weight given to sample $i$ by tree $t$ is $w_{i, t}^{DT}(x) = \mathbb{I}(x^i \in \mathcal{L}_t(x))/\mid \mathcal{L}_t(x) \mid$ where $\mathbb{I}(\cdot)$ is the indicator function and $\mathcal{L}_t(x)$ returns the leaf of tree $t$ that contains $x$. An important observation is that decision-tree predictors yield piecewise-constant decision policies since the weights are constant in each leaf. Consequently, random forests also define piecewise-constant decision policies.

Nearest-neighbor predictors assign equal weights to all $k$-nearest neighbors of the current context. The predictor weights are thus given by $w_{i}^{kNN} = \frac{1}{k} \mathbb{I} \left( x^i \in \mathcal{N}_k \left( x \right) \right)$, where $\mathcal{N}_k(x)$ is the set of $k$-nearest neighbors of $x$. Again, the weight functions given by a nearest-neighbor predictor are piecewise constant, and so is its prescribed policy.

\subsection{Contextual Stochastic Problems}
\paragraph{Problems with expected costs.}
Two-stage stochastic models allow the decision-maker to react to the observation of uncertain parameters by applying recourse decisions. As such, they form an important class of stochastic problems with widespread applications. Two-stage contextual stochastic problems with expected costs can be seen as a special case of Problem~\eqref{eq:wSAA} when the sample cost $c(z, y^i)$ is given by:
\begin{equation}
    c(z, y^i) = \frac{1}{n} c_1(z_1) + c_2(z_2^i ; z_1, y^i),
\end{equation}
where $c_1(z_1)$ is the immediate cost of the first-stage decision $z_1$ and $c_2(z_2^i ; z_1, y^i)$ is the second-stage cost of applying the recourse decision $z_2^i$.

\paragraph{Problems with Conditional Value-at-Risk}
Risk-averse problems reflect the risk preference of the decision-maker through risk measures. In particular, stochastic problems minimizing the conditional value-at-risk~(CVaR) are popular in a wide array of applications such as finance and energy management thanks to their advantageous mathematical properties \citep{Rockafellar2000}. Further, introducing risk considerations in data-driven optimization has an effect akin to regularization as it can decrease both the risk and expected costs \citep{Lin2022}.

The CVaR is defined as the average of the $\alpha$-tail of the cost distributions, where $\alpha$ represents the risk tolerance of the decision-maker \citep{Rockafellar2002}. Denote by $[n]$ the set of integer from $1$ to $n$ and let $\tau(\cdot;z): [n] \rightarrow [n]$ be a permutation that sorts the sample losses $\left(c(z, y^{\tau(1;z)}), \dots, c(z, y^{\tau(i;z)}), \dots, c(z, y^{\tau(n;z)})\right)$ in decreasing order. We use this permutation to identify the index $i_\alpha$ such that $\sum_{i=1}^{i_\alpha} w_{\tau(i;z)} \ge 1-\alpha$ and $\sum_{i=1}^{i_\alpha-1} w_{\tau(i;z)} < 1-\alpha$. The index $i_\alpha$ is unique and denotes the sample loss at which the $\alpha$-tail of the loss distribution is covered. For any decision $z$, the CVaR is thus given by:
\begin{equation}
    \label{eq:CVaR}
    \begin{split}
        \CVaR &(z ; w) =  \frac{1}{1-\alpha} \bigg(  \sum_{i=1}^{i_\alpha-1} w_{\tau(i;z)}
        c(z, y^{\tau(i;z)}) \\ + & \big(1-\alpha-\sum_{i=1}^{i_\alpha-1} w_{\tau(i;z)} \big) c(z, y^{\tau(i_\alpha;z)}) \bigg).
    \end{split}
\end{equation}

\subsection{Explaining Decisions through Alternative Contexts}
Let $x_{n+1}$ be the context in which we obtain the decision $z^*$, and let $\zAlt$ be an alternative decision. Consider now the general contextual stochastic problem given by:
\begin{equation}
    \label{eq:genCSO}
    z^* = \argmin_{z \in \mathcal{Z}} g\left(w(x_{n+1}), z\right),
\end{equation}
where $g$ is a cost function representing either the expected costs as in Problem~\eqref{eq:wSAA} or the CVaR defined in Equation~\eqref{eq:CVaR}. We introduce two classes of explanations for this general decision problem: relative and absolute explanations.
\begin{definition}[Relative explanation]
    \label{def:soft}
    A relative explanation $x^r$ is a context for which the alternative decision has a lower cost than the prescribed decision, that is:
    \begin{equation}
        g(w(x^r), \zAlt) \le g\left(w(x^r), z^*\right).
    \end{equation}
\end{definition}
\begin{definition}[Absolute explanation]
    \label{def:hard}
    An absolute explanation $x^a$ is a context for which the alternative decision is optimal, that is, $x^a$ satisfies the criterion:
    \begin{equation}
        \zAlt \in \argmin_{z \in \mathcal{Z}} g\left(w(x^a), z\right).
    \end{equation}
\end{definition}
Clearly, the absolute explanation criterion is stricter than the relative explanation criterion. In fact, the following property can be derived directly from Definitions~\ref{def:soft} and \ref{def:hard} as detailed in Appendix~\ref{app:proofs}.
\begin{proposition}
    \label{prop:absIsRel}
    An absolute explanation is always a relative explanation.
\end{proposition}

\paragraph{Existence of explanations.} In general, there is no guarantee that a relative or absolute explanation exists. The alternative decision might be poor and incur a large cost in all contexts. A direct consequence of Proposition~\ref{prop:absIsRel} is that, if no relative explanation exists, then no absolute explanation exists.

Sufficient conditions guaranteeing that no explanation exists can be identified from the historical samples alone. For instance, the following property holds for any predictor. The proof is given in Appendix~\ref{app:proofs}.
\begin{proposition}
    \label{prop:noExp}
    If $ c(\zAlt, y^i) \ge c(z^*, y^i)$ for all $i$, then there exists no relative or absolute explanation.
\end{proposition}

\textbf{Nearest explanations.} We are interested in finding relative and absolute explanations that are nearest to the current context $x_{n+1}$. Thus, nearest explanations minimize a distance function $f(\cdot, x_{n+1})$, and the following property can be deduced from Proposition~\ref{prop:absIsRel}.
\begin{corollary}
    \label{prop:softCloser}
    A nearest absolute explanation $x^a$ is at least as distant from the context $x_{n+1}$ as its nearest relative explanation $x^r$, i.e., $ f(x_a, x_{n+1}) \ge f(x_r, x_{n+1})$.
\end{corollary}
The proof is given in Appendix~\ref{app:proofs}.  Corollary~\ref{prop:softCloser} underlies the design of our algorithm to obtain absolute explanations presented in Section~\ref{sec:method}.

The above definitions and properties describe the structure of the explanation problem for both single-stage and two-stage stochastic programs with expected costs and CVaR objectives. Notice that, for two-stage problems, it is sufficient to explain first-stage decisions since second-stage decisions can be deduced from them. This is valuable in practice since there are often much fewer first-stage decisions than second-stage decisions.
\section{Obtaining Nearest Explanations}
\label{sec:method}
In the following, we develop methods to find nearest explanations based on integer programming. This approach is especially relevant since the decision policies are piecewise constant, and searching over the weights is a combinatorial problem. Moreover, integer programming solvers have made tremendous progress over the last decades, permitting nowadays to solve many problems of practical relevance at scale (e.g., searching for optimal counterfactual explanations of classes -- \citealt{Parmentier2021}).

\subsection{Explaining Problems with Expected Costs}
The relative explanation problem can be defined as:
\allowdisplaybreaks
\begin{subequations}
    \label{optim:relGen}
    \begin{alignat}{2}
        \min_{x \in \mathcal{X}}      & \quad && f(x, x_{n+1}) \label{optim:relGen:obj} \\
        \text{s.t.} &       && \sum\nolimits_{i=1}^n w_i(x) \delta^i(\zAlt, z^*) \le 0,\label{optim:relGen:w1}
    \end{alignat}
\end{subequations}
where $\delta^i(\zAlt, z^*) = c\left(\zAlt, y^i\right) - c\left(z^*, y^i\right)$ is the difference between the costs incurred by decisions $\zAlt$ and $z^*$ for the past observation $y^i$, which can be calculated offline. The key part of our methodology is thus to search over the weight functions $\{w_i(x)\}_{i=1}^n$ given by random forest and nearest-neighbor predictors.

\paragraph{Random forest predictors.} Our model extends the existing methods for counterfactual explanations in the classification setting based on integer programming. These models consist of a set of linear constraints that ensure that $x$ is in leaf node $v$ of tree $t$ if and only if the binary variable $l_{t,v}$ is equal to $1$. For the sake of conciseness, we do not restate these formulations and refer instead to \citet{Parmentier2021} for a detailed presentation.

Given variables $l_{t,v}$, nearest relative explanations can be obtained by introducing variables $w_{i}$ and constraints:
\begin{equation}
    w_{i} = \frac{1}{T} \sum_{t=1}^T \sum_{v \in \mathcal{V}_t^L} S_{t, v, i} \cdot l_{t,v}, \quad \forall i \in [n]
\end{equation}
where $S_{t, v, i}$ is the weight of sample $i$ in node $v$ of tree $t$, which can be calculated offline as $1/\mathcal{L}_t(x_i)$ if sample $i$ is in node $v$ of tree $t$ and $0$ otherwise. Note that the additional variables are all continuous so that the complexity of the problem is similar to the one of counterfactual explanations in the classification setting. This approach is flexible and could be combined with any counterfactual explanation method developed in the classification setting.

\paragraph{Nearest-neighbor predictors.} Contrary to random forests and to the best of our knowledge, there is no existing optimization model to obtain nearest explanations of nearest-neighbor classifiers. Hence, we develop a new formulation to track the set $\mathcal{N}_k$ of $k$ nearest neighbors of an explanation. The formulation is based on auxiliary binary variables $\lambda_i$, equal to $1$ if sample $i$ is in $\mathcal{N}_k$, and on a free variable $d$ that takes for value the radius of a disk around $x$ that contains all its $k$-nearest neighbors but not its $(k+1)$-nearest neighbors. The relative explanation of nearest-neighbor predictors can be modeled with the following constraints:
\allowdisplaybreaks
\begin{subequations}
    \label{optim:knn}
    \begin{alignat}{2}
                    &       && w_i = \lambda_i / k, \quad \forall i \in [n] \, , \label{optim:knn:w2} \\
                    &       && d_i = f(x, x^i) , \quad \forall i \in [n] \, , \label{optim:knn:c1} \\
                    &       && d_i \le d + M(1-\lambda_i), \quad \forall i \in [n] \label{optim:knn:c2}  \, ,\\
                    &       && d_i \ge d + \varepsilon - M \lambda_i, \quad \forall i \in [n] \label{optim:knn:c3}  \, ,\\
                    &       && \sum\nolimits_{i=1}^n \lambda_i = k. \label{optim:knn:c4}
    \end{alignat}
\end{subequations}
Constraint~\eqref{optim:knn:w2} assigns equal weights to all $k$ nearest neighbors. Constraint~\eqref{optim:knn:c1} tracks the distance between the historical samples and the explanation. Constraints~\eqref{optim:knn:c2}, \eqref{optim:knn:c3} and \eqref{optim:knn:c4} identify the $k$-nearest neighbors of $x$ through so-called "big-M" constraints, where $M$ is an upper bound on the distance $d_i$ and depends on the feature space.

\paragraph{Absolute explanations.}
The absolute explanation problem can be stated in a general form by replacing Constraint~\eqref{optim:relGen:w1} with the absolute criterion from Definition~\ref{def:hard}. This leads to a complex bi-level formulation and is notoriously hard to solve \citep{Kleinert2021}. However, when the underlying decision problem is linear, the absolute explanation problem can be formulated as a compact, single-level optimization problem. Let $\mathcal{Z}$ be a convex polytope $\mathcal{Z} = \{ z : A z \le b \}$ with $b\in \mathbb{R}^{d_b}$ and $A \in \mathbb{R}^{d_b \times d_z}$, and let the objective function in~\eqref{eq:wSAA} be expressed as $ \sum_{i=1}^n w_i c(z, y^i) = (\sum_{i=1}^n w_i d_i )^\top z $ with $d_i \in \mathbb{R}^{d_z}$.
\begin{proposition}
    \label{prop:absLin}
    When Problem~\eqref{eq:wSAA} is linear, the absolute explanation criterion can be integrated into Problem~\eqref{optim:relGen} by adding free variables $\mu \in \mathbb{R}^{d_b}$ and $(d_z + 1)$ linear constraints:
    \begin{align*}
            \left( \sum\nolimits_{i=1}^n w_i d_i 
 \right)^\top \zAlt \le b^\top \mu \, , \\
            A^\top \mu \le \sum\nolimits_{i=1}^n w_i d_i .
    \end{align*}
\end{proposition}
The proof follows from strong duality and is given in Appendix~\ref{app:proofs}. The added constraints act as certificates for global optimality. Other single-level reformulations could be identified when the decision problem has a special structure (e.g., using the Karush–Kuhn–Tucker conditions for convex decision problems).

Still, we are interested in explaining general decision problems that are not necessarily linear. Thus, we suggest another more general algorithm that provides absolute explanations regardless of the nature of the underlying decision problem. The algorithm is based on repeatedly solving the simpler relative explanation problem to obtain intermediate contexts $x^{(j)}$ and checking whether the solution satisfies the absolute explanation criterion. If this is not the case, the region of the feature space containing $x^{(j)}$ is cut from the feasible domain. This can be done by adding the following constraint for random forest predictors:
\begin{equation}
        \sum\nolimits_{t=1}^T l_{t, v_{t(j)}} \le (T-1), \label{eq:hardRf1}
\end{equation}
where $v_{t(j)}$ is the index of the leaf node of tree $t$ that contains $x^{(j)}$. For the $k$-nearest neighbor predictor, it is done by adding the constraint:
\begin{equation}
        \sum_{i \in \mathcal{N}_k(x^{(j)})} \lambda_i \le k-1.
\end{equation}
A valid inequality can also be added to tighten the search space at each iteration:
\begin{equation}
    \label{eq:VI}
    \sum\nolimits_{i=1}^n w_i (x^{(j)}) \delta^i(\zAlt, z^{*}_j) \le 0.
\end{equation}
The enriched relative explanation problem with added constraints is then solved again. The procedure is outlined in Algorithm~\ref{alg:absCf}.

\begin{algorithm}[htb]
    \caption{Iterative procedure for absolute explanations}
    \label{alg:absCf}
    \begin{algorithmic}
        \STATE {\bfseries Initialize:} $j=0$, $x^{(0)} = x_{n+1}$
        \WHILE{$x^{(j)}$ is not an absolute explanation}
        \STATE $j=j+1$
        \STATE Solve the enriched relative problem to obtain $x^{(j)}$.
        \STATE Solve Problem~\eqref{eq:wSAA} with context $x^{(j)}$ to obtain $z^{*}_j$.
        \IF{$\sum_{i=1}^n w_i \left(x^{(j)}\right) \delta^i(\zAlt, z^{*}_j) > 0$}
        \STATE Cut the region containing $x^{(j)}$.
        \STATE Tighten the search space by adding Constraint~\eqref{eq:VI}.
        \ELSE
        \STATE {\bfseries Return:} $x^a = x^{(j)}$
        \ENDIF
        \ENDWHILE
    \end{algorithmic}
\end{algorithm}

A more efficient implementation of Algorithm~\ref{alg:absCf} can be obtained by leveraging the capabilities of modern integer programming solvers. Using callbacks, Problem~\eqref{eq:wSAA} can be solved each time a feasible integer solution is found. The temporary solution can then be excluded from the search space using so-called ``lazy constraints`` if it does not satisfy the absolute explanation criterion, and valid inequality~\eqref{eq:VI} can be added. This avoids solving the relative explanation problem from scratch at each iteration and substantially reduces computation times.

\subsection{Explaining Problems with CVaR}
\label{sec:cvarFormulation}
Explaining problems with CVaR objectives is significantly more complex than problems with expected costs. Indeed, the algorithm must simultaneously identify the indices of the samples with the largest losses and track their weights. However, this apparently non-linear problem can be equivalently stated as a mixed-integer linear problem. The main idea is to express $\CVaR(\zAlt)$ and $\CVaR(z^*)$ using auxiliary variables and constraints and to introduce a constraint on the relative explanation criterion: $\CVaR(\zAlt) \le \CVaR(z^*)$.

\paragraph{Modeling $\CVaR(\zAlt)$.} We develop a flow-based formulation that relies on the continuous flow variables $\{f_i\}_{i=1}^n$ to track the weight given to each sample loss. The largest losses are identified using the binary variables $\{\theta_i\}_{i=1}^n$. The CVaR of the alternative decision $\zAlt$ can be expressed as:
\begin{equation}
    \label{eq:flowCvar}
    \CVaR(\zAlt) =\frac{1}{1-\alpha} \sum_{i=1}^n f_i \cdot c(\zAlt;y^{\tau(i, \zAlt)})
\end{equation}
by introducing the following constraints:
\begin{subequations}
    \begin{align}
        & \sum\nolimits_{i=1}^n f_i = 1-\alpha,\label{optim:cvar:c1} & \\
        & 0 \le f_i \le w_{\tau(i, \zAlt)}, & \forall i \in [n],\label{optim:cvar:c6}\\
        & \theta_i \ge \theta_{i+1}, & \forall i \in [n-1],\label{optim:cvar:c2} \\
        & w_{\tau(i, \zAlt)} - W_{\text{max}} (1-\theta_i) \le f_i, & \forall i \in [n],\label{optim:cvar:c3} \\
        & f_i \le W_{\text{max}} \theta_{i-1}, & \forall i \in [n],\label{optim:cvar:c4} \\
        & \sum\nolimits_{j=1}^i f_j \le \sum\nolimits_{j=1}^{i+1} f_j & \forall i \in [n-1]. \label{optim:cvar:c7}
    \end{align}
\end{subequations}
Constraint~\eqref{optim:cvar:c1} states that the sum of the flow variables covers the $\alpha$-tail of the loss distribution. Constraints~\eqref{optim:cvar:c6} ensures that the flows are always smaller than the sample weights. Constraint~\eqref{optim:cvar:c2} orders the auxiliary binary variables. Constraints~\eqref{optim:cvar:c3} and \eqref{optim:cvar:c4} link the flow variables to the binary variables using a big-M method. Thus, the flow of any sample $i$ such that $\tau(i, \zAlt) < \tau(i_\alpha, \zAlt)$ is exactly $w_{\tau(i, \zAlt)}$, and the flow of any samples $i$ such that $\tau(i, \zAlt) > \tau(i_\alpha, \zAlt)$ is exactly $0$. The flow of $i_\alpha$, the last sample selected is free between $0$ and $w_{\tau(i_\alpha, \zAlt)}$. These modeling steps are necessary since the CVaR calculation needs to ``split the atom'' with discrete distributions  (see \citet{Rockafellar2002}). We discuss how to derive tight upper bounds for $W_{\text{max}}$ in Appendix~\ref{app:cvarBounds}. Constraint~\eqref{optim:cvar:c7} is a valid inequality that tightens the feasible domain of the flow variables.

\paragraph{Efficient formulation for $\CVaR(z^*)$.} Let $\{\tilde{f}_i\}_{i=1}^n$ and $\{\tilde{\theta}_i\}_{i=1}^n$ be the additional variables introduced to track the continuous flow and binary indices of the largest sample losses for calculating $\CVaR(z^*)$, respectively. To model $\CVaR(z^*)$, the same set of constraints \eqref{eq:flowCvar}-\eqref{optim:cvar:c7} could be added. However, a more efficient approach can be used: it is sufficient to introduce only the continuous flow variables, as stated in the following proposition.
\begin{proposition}
    \label{prop:relCvar}
    The relative explanation criterion is satisfied if $\CVaR(\zAlt) \le \frac{1}{1-\alpha} \sum_{i=1}^n \tilde{f}_i \cdot c(z^*;y^{\tau(i, z^*)})$ and $\{\tilde{f}_i\}_{i=1}^n$ satisfy the inequalities~\eqref{optim:cvar:c1} and \eqref{optim:cvar:c6}.
\end{proposition}
The proof and the resulting formulation of the CVaR explanation problem are given in Appendix~\ref{app:proofs}. Proposition~\ref{prop:relCvar} implies that we do not need to introduce additional binary variables $\{\tilde{\theta}_i\}_{i=1}^n$, which reduces significantly the complexity of the explanation problem. The sample weights can then be modeled exactly as when explaining problems with expected costs, and Algorithm~\ref{alg:absCf} can be used to obtain absolute explanations.
\section{Numerical Study}
\label{sec:numStudy}
We demonstrate the value of our methods to explain data-driven decisions in repeated experiments with synthetic and real-world data. The goals of the experiments are threefold. First, we measure the scalability of our methods to different dimensions that impact the explanation, such as the number of training samples or decisions. Second, we investigate the characteristics and relative strengths of the two types of explanations introduced. Third, we present a practical application of our method to a risk-averse shortest-path problem based on Uber movement data.

All experiments are run on four cores of an Intel(R) Xeon(R) Gold 6336Y CPU at 2.40 GHz. The simulations are implemented in Python~3.9.13. Gurobi~9.5.1 is used to solve all mixed-integer programming problems. The code used to generate all the results in this paper is publicly available at \url{https://github.com/alexforel/Explainable-CSO} under an MIT license.

The random forest predictors are trained following \citet{Bertsimas2020}. We use the standard procedure of \textit{scikit-learn} v.1.0.2. with $T=100$ trees (default value) and a maximum depth of $4$. We focus on nearest-neighbor predictors with $k=10$. All distances are measured through the $l_1$ norm, which encourages sparse explanations in which only a few features are modified.

\subsection{Experimental Setting}
We select three stochastic decision problems that cover essential applications in operations management such as inventory management and routing:
\begin{itemize}[topsep=-0.5\parskip, itemsep=3pt, wide=5pt]
    \item \textbf{Multi-item newsvendor.} A newsvendor manages several products and wants to optimize profits in the face of uncertain demands and a limited ordering budget. This problem is adapted from \citet{Kallus2022}.
    \item \textbf{Two-stage shipment planning.} The decision-maker aims to minimize production at several facilities (first stage) and shipping costs (second stage) while satisfying the uncertain demand at several locations. This problem is taken from \citet{Bertsimas2020}.
    \item \textbf{(CVaR) shortest path.} The decision-maker seeks the shortest path that traverses a network with uncertain edge costs. The problem is adapted from \citet{Elmachtoub2020} and \citet{Elmachtoub2022}.
\end{itemize}

An overview of the problems' dimensions is given in Table~\ref{tab:expDim} and detailed descriptions are provided in Appendix~\ref{app:exp}. In our experiments, we analyze different configurations of the decision problems and perform $N=100$ repeated experiments in each setting.  The training data $\{(x^i, y^i)\}_{i=1}^n$, corresponding to historical observations of the contextual information and the random parameters of the decision problem, is resampled in each experiment.
\begin{table}[ht]
    \vspace{-.3cm}
    \caption{Dimension of contexts, observations, and decisions.}
    \label{tab:expDim}
    \centering
    \begin{sc}
        \resizebox{0.87\linewidth}{!}{\begin{tabular}{lccc}
    \toprule
    \textbf{Problem} & $d_x$   & $d_y$ & $d_z$ \\
    \midrule
    Newsvendor (NWS)       & $2$     & $20$  & $20$ \\
    Shipment (SHM)         & $3$     & $12$  & $8+48n$\\
    Shortest path (SP)    & $4$     & $112$  & $112$ \\
    CVaR shortest-path (c-SP)       & $4$     & $24$   &  $24$ \\
    \bottomrule
\end{tabular}}
        \vspace{-.3cm}
    \end{sc}
\end{table}

\paragraph{Generating explanation problems.} To generate explanation problems, we need an initial context $x_{n+1}$, its associated decision $z^*$, and an alternative decision $\zAlt$ to explain. We create diverse explanation problems by sampling two contexts. The first one is taken as the initial context $x_{n+1}$ and its associated decision $z^*$ is obtained by solving Problem~\eqref{eq:wSAA}. The second context $\xAlt$ is used to determine an alternative decision $\zAlt$, also by solving Problem~\eqref{eq:wSAA}. We then obtain relative and absolute explanations by solving Problem~\eqref{optim:relGen} and applying Algorithm~\ref{alg:absCf} respectively.

We present below the main results of our experiments. Additional results, such as a sensitivity analysis of the model hyper-parameters or measuring the distance of the obtained explanations, are presented in Appendix~\ref{app:numRes}.
 
\subsection{Numerical Results on Synthetic Experiments}
\label{sec:numRes}
First, we measure the time needed to obtain relative and absolute explanations. The average time is given in Table~\ref{tab:compTimes} for varying sizes of the training set $\{(x^i, y^i)\}_{i=1}^n$. Overall, explanations can be obtained in times ranging from a few seconds to a few minutes. Our methods scale especially well with the size of the training set when the predictor is a random forest. Explanations of nearest-neighbor predictors can be obtained in short times for small sample sizes, but do not scale as well to large sample sizes. For both predictors, explaining problems optimized over CVaR objectives is more computationally challenging than those optimized over expected costs.
\begin{table}[ht]
    \vspace{-.3cm}
    \caption{Average computational time [in s].}
    \label{tab:compTimes}
    \centering
    \begin{sc}
        \resizebox{\linewidth}{!}{\begin{tabular}{*{10}{c}} 
    \toprule   &  & \multicolumn{2}{c}{Relative explanation} & \multicolumn{2}{c}{Absolute explanation} \\ 
    \cmidrule(lr){3-4} \cmidrule(lr){5-6} & n & RF & k-NN & RF & k-NN \\ 
    \midrule \multirow{3}{*}{NWS} & $50$ & 0.28 & 0.24 & 1.65 & 0.96 \\ 
     & $100$ & 0.37 & 1.07 & 3.00 & 5.46 \\ 
     & $200$ & 0.48 & 3.71 & 4.41 & 23.90 \\ 
    \cmidrule{2-6} \multirow{3}{*}{SHM} & $50$ & 0.34 & 0.25 & 2.68 & 2.06 \\ 
     & $100$ & 0.36 & 1.10 & 5.61 & 9.25 \\ 
     & $200$ & 0.47 & 4.12 & 14.35 & 53.32 \\ 
    \cmidrule{2-6} \multirow{3}{*}{SP} & $50$ & 0.61 & 0.87 & 1.78 & 2.66 \\ 
     & $100$ & 1.02 & 5.34 & 3.67 & 33.77 \\ 
     & $200$ & 1.41 & 24.25 & 5.50 & 141.53 \\ 
    \cmidrule{2-6} \multirow{3}{*}{c-SP} & $50$ & 6.78 & 1.40 & 7.47 & 4.47 \\ 
     & $100$ & 16.27 & 9.45 & 17.88 & 25.60 \\ 
     & $200$ & 45.20 & 37.66 & 48.63 & 203.52 \\ 
\bottomrule 
\end{tabular}}
        \vspace{-.1cm}
    \end{sc}
\end{table}

\paragraph{Sensitivity to the dimension of the context.} We evaluate the scalability of explanation methods for random forest predictors in both the sample size and the number of features through additional sensitivity analyses. We focus on the shortest path problem (SP) and vary the dimension of the contextual information $d_x \in \{5, 10, 25, 50, 100, 500 \}$ with training sets of different sizes. In this experiment, all the features are informative and may influence the uncertain travel times on each edge of the network. Results on the CVaR shortest-path problem (c-SP) are presented in Appendix~\ref{app:addResults}.
\begin{figure}[ht!]
    \centering
    \vspace*{-1mm}
    \resizebox{\linewidth}{!}{\tikzset{dgStyle/.style={darkgreen, mark=triangle*,mark options={fill=darkgreen}}}

\begin{tikzpicture}
	\begin{groupplot}[
		group style={
			group name=my plots,
			group size=2 by 1,
			xlabels at=edge bottom,
			ylabels at=edge left
		},
		height = 4.5cm,
		width  = 5.5cm,
        ymode=log,
        xmin=5, xmax=500,
        xmode=log,
        xtick = {5, 10, 25, 50, 100, 500},
        xticklabels = {5, 10, 25, 50, 100, 500},
        ytick = {0.1, 1, 10, 100},
        yticklabels = {0.1, 1, 10, 100},
		xlabel = {Context dimension $d_x$},
		ylabel = {Time [in s]},
        ylabel shift = -2 pt,
        xlabel shift = -2 pt
		]
		\nextgroupplot[title = {(a) Relative explanations}, title style={yshift=-5pt},
                        font = \small]
        %%%%%%%%%%% n=100 %%%%%%%%%%%
		\addplot+[blue] table [x index = {0}, y index = {1}, col sep=comma]{plots/path_feat_sens.csv};
        \addplot+[name path=C, blue!20, mark=none, forget plot] table [x index = {0}, y index = {2}, col sep=comma]{plots/path_feat_sens.csv};
        \addplot+[name path=D, blue!20, mark=none, forget plot] table [x index = {0}, y index = {3}, col sep=comma]{plots/path_feat_sens.csv};
        \addplot[blue, fill opacity=0.05, forget plot] fill between[of=C and D];
        %%%%%%%%%%% n=500 %%%%%%%%%%%
		\addplot+[red] table [x index = {0}, y index = {7}, col sep=comma]{plots/path_feat_sens.csv};
        \addplot+[name path=C, red!20, mark=none, forget plot] table [x index = {0}, y index = {8}, col sep=comma]{plots/path_feat_sens.csv};
        \addplot+[name path=D, red!20, mark=none, forget plot] table [x index = {0}, y index = {9}, col sep=comma]{plots/path_feat_sens.csv};
        \addplot[red, fill opacity=0.05, forget plot] fill between[of=C and D];
        %%%%%%%%%%% n=1000 %%%%%%%%%%%
		\addplot+[dgStyle] table [x index = {0}, y index = {13}, col sep=comma]{plots/path_feat_sens.csv};
        \addplot+[name path=C, darkgreen!20, mark=none, forget plot] table [x index = {0}, y index = {14}, col sep=comma]{plots/path_feat_sens.csv};
        \addplot+[name path=D, darkgreen!20, mark=none, forget plot] table [x index = {0}, y index = {15}, col sep=comma]{plots/path_feat_sens.csv};
        \addplot[darkgreen, fill opacity=0.05, forget plot] fill between[of=C and D];
        %%%%%%%%%%% n=5000 %%%%%%%%%%%
		\addplot+[black] table [x index = {0}, y index = {19}, col sep=comma]{plots/path_feat_sens.csv};
        \addplot+[name path=C, black!20, mark=none, forget plot] table [x index = {0}, y index = {20}, col sep=comma]{plots/path_feat_sens.csv};
        \addplot+[name path=D, black!20, mark=none, forget plot] table [x index = {0}, y index = {21}, col sep=comma]{plots/path_feat_sens.csv};
        \addplot[black, fill opacity=0.05, forget plot] fill between[of=C and D];
		
		\nextgroupplot[title = {(b) Absolute explanations}, title style={yshift=-5pt}, font = \small,
			legend style={at={(-0.125,-0.35)},anchor=north},
			legend columns=4]
        %%%%%%%%%%% n=100 %%%%%%%%%%%
		\addplot+[blue] table [x index = {0}, y index = {4}, col sep=comma]{plots/path_feat_sens.csv};
        \addlegendentry{$n=100$}
        \addplot+[name path=C, blue!20, mark=none, forget plot] table [x index = {0}, y index = {5}, col sep=comma]{plots/path_feat_sens.csv};
        \addplot+[name path=D, blue!20, mark=none, forget plot] table [x index = {0}, y index = {6}, col sep=comma]{plots/path_feat_sens.csv};
        \addplot[blue, fill opacity=0.05, forget plot] fill between[of=C and D];
        %%%%%%%%%%% n=500 %%%%%%%%%%%
		\addplot+[red] table [x index = {0}, y index = {10}, col sep=comma]{plots/path_feat_sens.csv};
        \addlegendentry{$n=500$}
        \addplot+[name path=C, red!20, mark=none, forget plot] table [x index = {0}, y index = {11}, col sep=comma]{plots/path_feat_sens.csv};
        \addplot+[name path=D, red!20, mark=none, forget plot] table [x index = {0}, y index = {12}, col sep=comma]{plots/path_feat_sens.csv};
        \addplot[red, fill opacity=0.05, forget plot] fill between[of=C and D];
        %%%%%%%%%%% n=1000 %%%%%%%%%%%
		\addplot+[dgStyle] table [x index = {0}, y index = {16}, col sep=comma]{plots/path_feat_sens.csv};
        \addlegendentry{$n=1000$}
        \addplot+[name path=C, darkgreen!20, mark=none, forget plot] table [x index = {0}, y index = {17}, col sep=comma]{plots/path_feat_sens.csv};
        \addplot+[name path=D, darkgreen!20, mark=none, forget plot] table [x index = {0}, y index = {18}, col sep=comma]{plots/path_feat_sens.csv};
        \addplot[darkgreen, fill opacity=0.05, forget plot] fill between[of=C and D];
        %%%%%%%%%%% n=5000 %%%%%%%%%%%
		\addplot+[black] table [x index = {0}, y index = {22}, col sep=comma]{plots/path_feat_sens.csv};
        \addlegendentry{$n=5000$}
        \addplot+[name path=C, black!20, mark=none, forget plot] table [x index = {0}, y index = {23}, col sep=comma]{plots/path_feat_sens.csv};
        \addplot+[name path=D, black!20, mark=none, forget plot] table [x index = {0}, y index = {24}, col sep=comma]{plots/path_feat_sens.csv};
        \addplot[black, fill opacity=0.05, forget plot] fill between[of=C and D];
	\end{groupplot}
\end{tikzpicture}}
    \vspace*{-4mm}
    \caption{SP: computational time on large instances (standard deviation is shown in shaded area).}
    \label{fig:featSens}
    \vspace*{-2mm}
\end{figure}
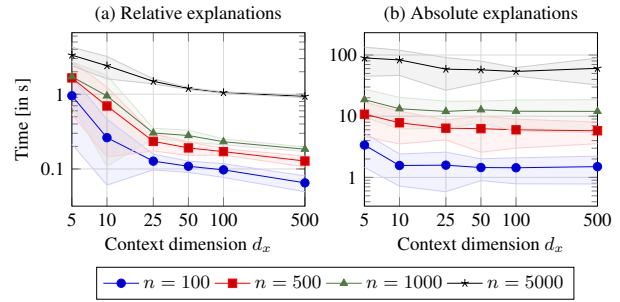 

We show the average time to find explanations in Figure~\ref{fig:featSens}, where the shaded areas correspond to the interval of plus/minus one standard deviation. Figure~\ref{fig:featSens} shows that explanations of random forest predictors can be obtained in a short time even when the dimension of the context or the size of the training set is large. Interestingly, the computational effort of relative explanations decreases as the dimension of the context increases.

In the RF model, the number of features only impacts the number of continuous variables of the problem. We observe in our experiments that the linear relaxation of the underlying problem becomes stronger as the dimension of the context increases. As a consequence, the number of branch-and-bound nodes and the solution time of the mixed-inter optimization problem tend to decrease.
 
\begin{figure}[ht]
	\centering
	\includegraphics[trim=2mm 2mm 2mm 2mm, clip, width=0.95\linewidth]{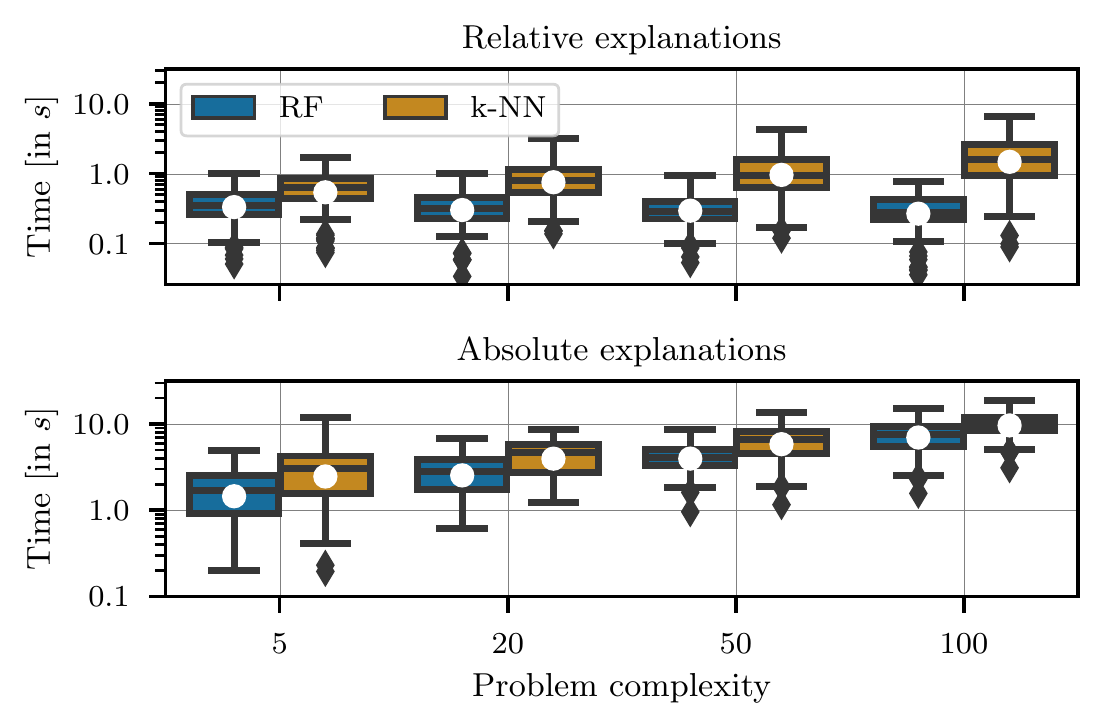}
    \vspace*{-2mm}
	\caption{NWS: sensitivity to the number of products.}
	\label{fig:probSensNews}
    \vspace{-.4cm}
\end{figure}

\begin{figure}[ht]
	\centering
	\includegraphics[trim=2mm 2mm 2mm 2mm, clip, width=0.95\linewidth]{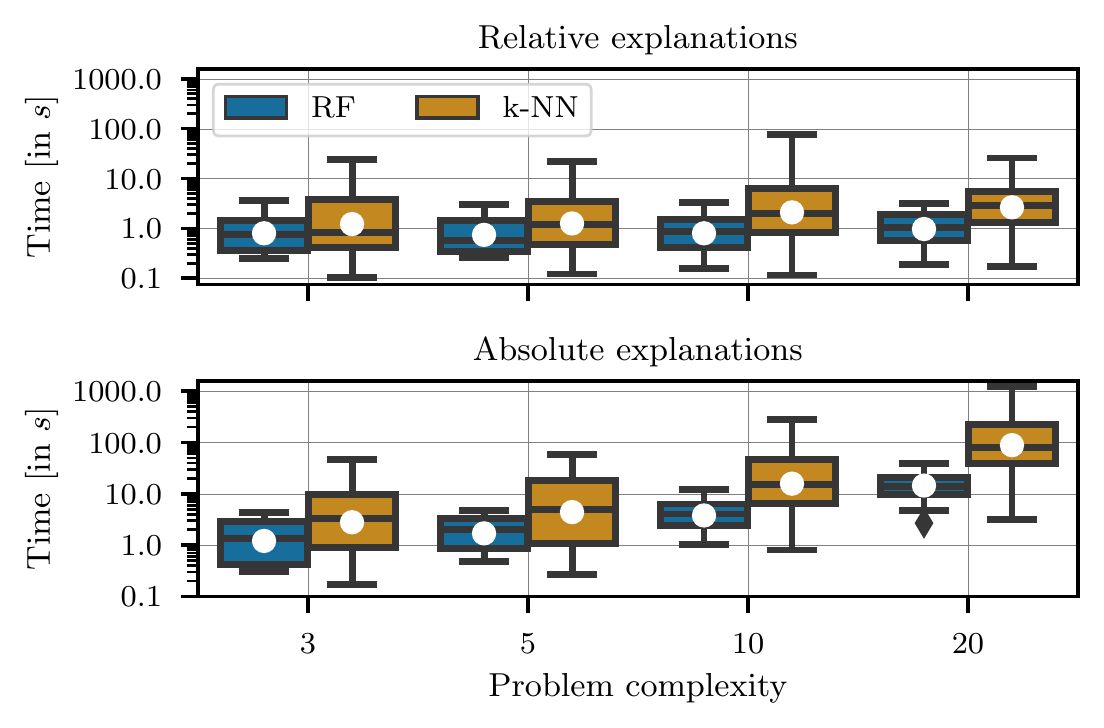}
    \vspace*{-2mm}
	\caption{SP: sensitivity to the grid width.}
	\label{fig:probSensPath}
    \vspace{-.4cm}
\end{figure}

\paragraph{Sensitivity to the decision problem.} We now investigate how the complexity of the wSAA Problem~\eqref{eq:wSAA} impacts the time needed to explain it. We vary the number of products in the newsvendor problem and the size of the grid in the shortest-path problem. Thus, we increase the dimensions $d_y$ and $d_z$ while keeping the dimension $d_x$ and the size of the training set constant ($n=100$). The time needed to find explanations for the newsvendor and shortest-path problem is shown in Figure~\ref{fig:probSensNews} and \ref{fig:probSensPath}, respectively. The results show that the computational time of relative explanations scales remarkably well with the complexity of the decision problem.

\begin{figure}[ht]
	\centering
    \vspace{-.1cm}
	\includegraphics[trim=0mm 0mm 0mm 0mm, clip, width=0.75\linewidth]{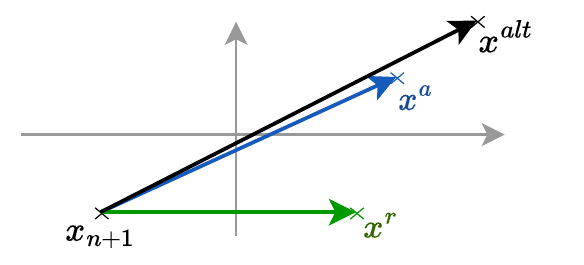}
    \vspace{-.2cm}
	\caption{Explanations $x^a$ and $x^r$ with large positive correlations with the alternative context $\xAlt$.}
	\label{fig:corrExample}
    \vspace{-.3cm}
\end{figure}

\paragraph{Identifying relevant feature changes.} In each experiment, the alternative decision $\zAlt$ is obtained from a sampled context $\xAlt$. We can measure whether our explanations move toward this alternative context, thus identifying the right feature changes that render the alternative decision $\zAlt$ better or optimal. We focus on the newsvendor problem since it does not present too much symmetry in the context (see Appendix~\ref{app:symSol}). We measure the correlation $c\in[-1,1]$ between the relative explanation $x^r$ produced by our method (or the absolute explanation $x^a$) and the alternative context $\xAlt$ as a normalized dot product: $c = \frac{\xAlt - x_{n+1}}{\lVert \xAlt - x_{n+1} \rVert_2} \cdot \frac{x^r - x_{n+1}}{\lVert x^r - x_{n+1} \rVert_2} $. A large positive correlation indicates that the two contexts are in the same direction, as shown in Figure~\ref{fig:corrExample}.

\begin{figure}[ht]
    \vspace{-.1cm}
    \centering
    \includegraphics[trim=2mm 2mm 2mm 2mm, clip, width=0.98\linewidth]{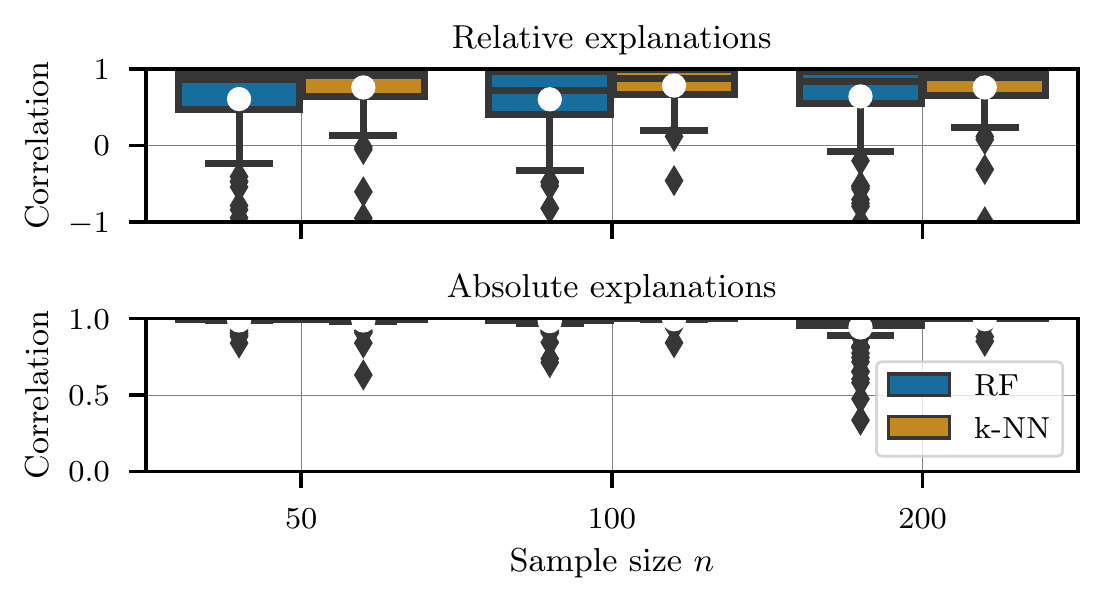}
    \vspace*{-2mm}
    \caption{NWS: Correlation between explanations and $\xAlt$.}
    \label{fig:newsCorr}
    \vspace{-.2cm}
\end{figure}

The correlations between our explanations and the alternative contexts are shown in Figure~\ref{fig:newsCorr} for training sets of varying sizes. As seen in this experiment, the explanations produced by our algorithm are aligned with the alternative contexts. In fact, the absolute explanations and alternative contexts are almost perfectly correlated. Relative explanations also have a largely positive correlation, although lower than absolute explanations. A possible interpretation is that the relative explanation criterion allows deviating from the direction of the alternative contexts if it leads to sparser explanations, as encouraged by our use of the $l_1$ distance.

In Appendix~\ref{app:spurious}, we extend this experiment by including spurious features. The results highlight that relative explanations tend to focus on relevant features and avoid spurious ones, whereas absolute explanations always have a high correlation with the alternative contexts regardless of the importance of the features used in the explanation.

\subsection{CVaR Shortest Path with Real-World Data}
\label{sec:uberPath}
We now detail the application of our methods to the CVaR shortest-path problem based on Uber movement data. The decision-maker seeks a path from the east to the west of the Los Angeles downtown area that minimizes a $\CVaR$ objective with $\alpha=0.7$. The road network is modeled as a graph of $45$ nodes and $93$ edges. The contextual information contains temporal as well as meteorological information. The '\textsc{Period}' feature, indicating the time of the day, is modeled as a categorical feature. The '\textsc{Day}' and '\textsc{Month}' features are discrete, and the remaining features are continuous. Because our method is based on integer programming, it is straightforward to handle discrete or one-hot encoded categorical features.

Historical observations of the uncertain travel times are available on the edge level. We randomly sample $n=1000$ past observations and train a random forest predictor. We then sample a new context $x_{n+1}$ and calculate its corresponding optimal path $z^*$. The solution is to traverse the city using the north route, shown in blue in Figure~\ref{fig:uberPath}. We consider the alternative route $\zAlt$ that takes instead the south route, shown in green in Figure~\ref{fig:uberPath}, and wish to identify in which context such an alternative solution would be relevant.

We can compute the nearest relative and absolute explanations in $43$ and $63$ seconds respectively. The initial context and its nearest explanations are given in Table~\ref{tab:uberExp}. The results show that the alternative route would be better, actually even optimal, if the '\textsc{Period}' feature changes from 'Midday' to 'AM'. In this example, the explanation is sparse and can be easily interpreted by decision-makers. For instance, a direct interpretation is that the traffic along the alternative path is higher around midday than in the early morning.
\begin{table}[ht]
    \vspace{-.2cm}
    \caption{Nearest explanations for the Uber shortest-path problem.}
    \label{tab:uberExp}
    \centering
    \begin{sc}
    \resizebox{\linewidth}{!}{
    \begin{tabular}{l*{7}{c}} 
    \toprule
        Feature & Initial $x_{n+1}$ & Relative $x^r$ & Absolute $x^a$ \\
    \midrule
        Period & \textbf{Midday} & \textbf{AM}  & \textbf{AM}  \\
        Temperature & 57.17 & 57.17  & 57.17  \\
        Wind speed & 4 & 4  & 4  \\
        Rain & 0 & 0  & 0  \\
        Visibility & 6.99 & 6.99  & 6.99  \\
        Day & 2 & 2  & 2  \\
        Month & 11 & 11  & 11 \\        
    \bottomrule
    \end{tabular}
    }
    \end{sc}
    \vspace{-.3cm}
\end{table}
\section{Conclusion and broader impact}
\label{sec:conclusion}
This paper presents a first approach to explain the decisions made by data-driven pipelines involving machine learning predictors. We obtained nearest explanations by analyzing the structure of the explanation problems and solving integer-programming models. Approximate explanations could also be investigated for complex cases such as problems with CVaR objectives, building upon the rich literature on approximate explanations in the classification setting \citep[see e.g.,][]{Tolomei2017, Lucic2022}. Our approach would then serve as a benchmark with provably minimal explanation distance.

This work located at the intersection of explainable AI and data-driven decision-making opens many promising research avenues. Future work could investigate how to identify a set of diverse but similar explanations, or consider novel types of explanations, such as finding the nearest context such that the current decision is not optimal. Another interesting research direction is to develop explanation techniques for the smart "predict-then-optimize" framework of \citet{Elmachtoub2020} or other integrated pipelines \citep{Dalle2022, Ferber2020, Mandi2020}.

\section*{Acknowledgements}
The authors want to thank Vahid Partovi Nia and Matthias Soppert for their helpful comments. We also acknowledge the support of IVADO and the Canada First Research Excellence Fund (Apog\'ee/CFREF).

\bibliography{bibliography}
\bibliographystyle{icml2023}

%%%%%%%%%%%%%%%%%%%%%%%%%%%%%%
%%%%%%%%%%%%%%%%%%%%%%%%%%%%%%
% APPENDIX
%%%%%%%%%%%%%%%%%%%%%%%%%%%%%%
%%%%%%%%%%%%%%%%%%%%%%%%%%%%%%
\newpage
\appendix
\onecolumn
\section{Supplementary Material: Proofs}
\label{app:proofs}
This section provides the proofs of the results presented in Section~\ref{sec:problem} and \ref{sec:method}. For the sake of completeness, we restate the result  before each proof.

%%%%%%%%%%%%%%%%%%%%%%%%%%%%%%%%%%%%%%%%
%%%%%%%%%%%%     PROOFS     %%%%%%%%%%%%
%%%%%%%%%%%%%%%%%%%%%%%%%%%%%%%%%%%%%%%%
\textbf{Proposition~\ref{prop:absIsRel}} An absolute explanation is always a relative explanation.
\begin{proof}
    The definition of absolute explanations implies that $g(x_a, \zAlt) \le g(x_a, z), \, \forall z \in \mathcal{Z}$. Hence, since $z^* \in \mathcal{Z}$, $x_a$~satisfies the relative explanation criterion.
\end{proof}

%%%%%%%%%%%%%%%%%%%%%%%%%%%%%%%%%%%%%%%%
\textbf{Proposition~\ref{prop:noExp}} If $ c(\zAlt, y^i) \ge c(z^*, y^i)$ for all $i$, then there exists no relative or absolute explanation.
\begin{proof}
    Let $\{w_i\}_{i=1}^n$ be an arbitrary set of weights such that $w_i \in [0,1], \, \forall i $ and $\sum_{i=1}^n w_i = 1$. First, consider decision problems minimizing expected costs. Since the weights $w_i$ are positive, $ c(\zAlt, y^i) \ge c(z^*, y^i)$ for all $i$ implies that $\sum_{i=1}^n w_i (c(\zAlt, y^i) - c(z^*, y^i)) \ge 0$. Thus no relative explanation exists and therefore no absolute explanation exists.
    
    Second, consider decision problems minimizing the CVaR. Let $\tau(\cdot ; z^*)$ be the permutation that orders all the sample losses $\{c(z^*, y^{\tau(i ; z^*)} \}_{i=1}^n$ in decreasing order and let $i_\alpha$ be the index at which the $\alpha$-tail of the loss distribution is covered. The following inequality follows from the fact that $ c(\zAlt, y^i) \ge c(z^*, y^i)$ for all $i$:
    \begin{equation}
        \label{eq:cvarIneq}
        \begin{split}
                \CVaR (z^* ; w) =  \frac{1}{1-\alpha} \bigg(  \sum_{i=1}^{i_\alpha-1} w_{\tau(i;z^*)}
                c(z^*, y^{\tau(i;z^*)}) +  \big(1-\alpha-\sum_{i=1}^{i_\alpha-1} w_{\tau(i;z^*)} \big) c(z^*, y^{\tau(i_\alpha;z^*)}) \bigg) \\
                \le \frac{1}{1-\alpha} \bigg(  \sum_{i=1}^{i_\alpha-1} w_{\tau(i;z^*)}
                c(\zAlt, y^{\tau(i;z^*)}) +  \big(1-\alpha-\sum_{i=1}^{i_\alpha-1} w_{\tau(i;z^*)} \big) c(\zAlt, y^{\tau(i_\alpha;z^*)}) \bigg).
        \end{split}
    \end{equation}
    Further, we know from the definition of the CVaR that:
    \begin{equation*}
        \frac{1}{1-\alpha} \bigg(  \sum_{i=1}^{i_\alpha-1} w_{\tau(i;z^*)}
                c(\zAlt, y^{\tau(i;z^*)}) +  \big(1-\alpha-\sum_{i=1}^{i_\alpha-1} w_{\tau(i;z^*)} \big) c(\zAlt, y^{\tau(i_\alpha;z^*)}) \bigg) \le \CVaR (\zAlt ; w).
    \end{equation*}
    Thus, no relative or absolute explanation exists.
\end{proof}

%%%%%%%%%%%%%%%%%%%%%%%%%%%%%%%%%%%%%%%%
\textbf{Corollary~\ref{prop:softCloser}} A nearest absolute explanation $x^a$ is at least as distant from the context $x_{n+1}$ as its nearest relative explanation $x^r$, i.e., $ f(x_a, x_{n+1}) \ge f(x_r, x_{n+1})$.
\begin{proof}
     The proof is by contradiction. If $x^a$ is a nearest absolute explanation and $f(x_a, x_{n+1}) < f(x_r, x_{n+1})$, then $x^a$ is a nearer relative explanation than $x^r$, which contradicts $x^r$ being a nearest relative explanation.
\end{proof}

%%%%%%%%%%%%%%%%%%%%%%%%%%%%%%%%%%%%%%%%
\textbf{Proposition~\ref{prop:absLin}} When Problem~\eqref{eq:wSAA} is linear, the absolute explanation criterion can be integrated into Problem~\eqref{optim:relGen} by adding free variables $\mu \in \mathbb{R}^{d_z}$ and $(d_z + 1)$ constraints:
    \begin{align*}
            \left( \sum\nolimits_{i=1}^n w_i d_i 
 \right)^\top \zAlt \le b^\top \mu \, , \\
            A^\top \mu \le \sum\nolimits_{i=1}^n w_i d_i .
    \end{align*}
\begin{proof}
    The proof is based on strong duality and borrows techniques from robust optimization \citep{BenTal2009}. Recall the notation: $\mathcal{Z} = \{ z : A z \le b \}$ with $b\in \mathbb{R}^{d_b}$ and $A \in \mathbb{R}^{d_b \times d_z}$, and the objective of the decision problem is given by $ (\sum_{i=1}^n w_i d_i )^\top z $ with $d_i \in \mathbb{R}^{d_z}$. The absolute explanation criterion $\zAlt \in \argmin_{z \in \mathcal{Z}} (\sum_{i=1}^n w_i d_i )^\top z$ can be equivalently expressed as:
    \begin{equation}
        \label{eq:absMin}
         \left(\sum_{i=1}^n w_i d_i \right)^\top \zAlt \le \min_{z \in \mathcal{Z}} \left( \sum_{i=1}^n w_i d_i \right)^\top z.
    \end{equation}
    By strong duality, since the minimization problem on the right-hand side of Equation~\eqref{eq:absMin} is feasible and bounded, its dual problem is also feasible and bounded and their optimal values coincide. The absolute explanation criterion is equivalent to:
    \begin{equation}
    \label{eq:absDual}
        \left(\sum_{i=1}^n w_i d_i \right)^\top \zAlt \le \max_{\mu \in \{\mu: A^\top \mu \le  \sum_{i=1}^n w_i d_i\}} b^\top \mu.
    \end{equation}
    Notice, however, that the dual problem on the right-hand side of Equation~\eqref{eq:absDual} does not need to be solved to optimality. Indeed, any feasible $\mu^0 \in \{\mu: A^\top \mu \le  \sum_{i=1}^n w_i d_i\}$ satisfies $b^\top \mu^0 \le \max_{\mu \in \{\mu: A^\top \mu \le  \sum_{i=1}^n w_i d_i\}} b^\top \mu$. Thus, the following condition is sufficient to ensure the absolute explanation criterion:
    \begin{equation*}
        \exists \, \mu \text{ such that }
        \begin{cases}
            \left(\sum_{i=1}^n w_i d_i \right)^\top \zAlt \le b^\top \mu, \text{ and } \\
            A^\top \mu \le  \sum_{i=1}^n w_i d_i.
        \end{cases}
    \end{equation*}
\end{proof}

%%%%%%%%%%%%%%%%%%%%%%%%%%%%%%%%%%%%%%%%
\textbf{Proposition~\ref{prop:relCvar}} The relative explanation criterion is satisfied if $\CVaR(\zAlt) \le \frac{1}{1-\alpha} \sum_{i=1}^n \tilde{f}_i \cdot c(z^*;y^{\tau(i, z^*)})$ and $\{\tilde{f}_i\}_{i=1}^n$ satisfy the inequalities~\eqref{optim:cvar:c1} and \eqref{optim:cvar:c6}.
\begin{proof}
    Let $\{ w_i\}_{i=1}^n$ be an arbitrary set of weights that sum to $1$ and denote by $\mathcal{F}$ the set of positive variables that satisfy the inequalities in constraints~\eqref{optim:cvar:c1} and \eqref{optim:cvar:c6}, that is $\mathcal{F} = \{ f \in [0, 1]^n, \sum\nolimits_{i=1}^n f_i = 1-\alpha \text{ and } 0 \le f_i \le w_{\tau(i, z)}, \, \forall i \in [n] \}$. The following inequality is true for any $ \tilde{f} = \{ \tilde{f}_i \}_{i=1}^n \in \mathcal{F}$:
    \begin{equation*}
        \frac{1}{1-\alpha} \sum_{i=1}^n \tilde{f}_i \cdot c(z^*;y^{\tau(i, z^*)}) \le \max_{\{f_i\} \in \mathcal{F}} \frac{1}{1-\alpha} \sum_{i=1}^n f_i \cdot c(z^*;y^{\tau(i, z^*)}) = \CVaR(z^*).
    \end{equation*}
    Thus, the following condition guarantees that the relative explanation criterion is satisfied:
    \begin{equation*}
    \exists \, \tilde{f}_i \in \mathcal{F} \text{ such that } \CVaR(\zAlt) \le \frac{1}{1-\alpha} \sum_{i=1}^n \tilde{f}_i \cdot c(z^*;y^{\tau(i, z^*)}).
    \end{equation*}
    The relative explanation problem with CVaR objectives can now be formulated efficiently as:
    \begin{subequations}
        \begin{align}
            \min_{x \in \mathcal{X}} & \quad && \ell_1(x, x_{n+1})  \\
            \text{s.t.} &       && \sum_{i=1}^n f_i \cdot c(\zAlt;y^{\tau(i, \zAlt)}) \le \sum_{i=1}^n \tilde{f}_i \cdot c(z^*;y^{\tau(i, z^*)}), \\
                        &       && \sum_{i=1}^n f_i = 1-\alpha, & \\
                        &       && f_i \le w_{\tau(i, \zAlt)}(x), & \forall i \in [n],\\
                        &       && \theta_i \ge \theta_{i+1}, & \forall i \in [n-1], \\
                        &       && w_{\tau(i, \zAlt)}(x) - W_{\text{max}} (1-\theta_i) \le f_i, & \forall i \in [n], \\
                        &       && f_i \le W_{\text{max}} \theta_{i-1}, & \forall i \in [n], \\
                        &       && \sum_{j=1}^i f_j \le \sum_{j=1}^{i+1} f_j & \forall i \in [n-1] \\
                        &       && \sum_{i=1}^n \tilde{f}_i = 1-\alpha, & \\
                        &       && \tilde{f}_i \le w_{\tau(i, z^*)}(x), & \forall i \in [n] \\
                        &       && f, \tilde{f}_i \ge 0, \quad \theta_{i} \in \{0, 1\} & \forall i \in [n].
        \end{align}
    \end{subequations}
\end{proof}

%%%%%%%%%%%%%%%%%%%%%%%%%%%%%%%%%%%%%%%%
%%%%%%%%%%%% END OF PROOFS  %%%%%%%%%%%%
%%%%%%%%%%%%%%%%%%%%%%%%%%%%%%%%%%%%%%%%
\section{Supplementary Material: Model Refinements}
\label{app:cvarBounds}
\begin{figure}[ht]
    \centering
    \includegraphics[trim=0mm 0mm 0mm 0mm, clip, width=0.75\linewidth]{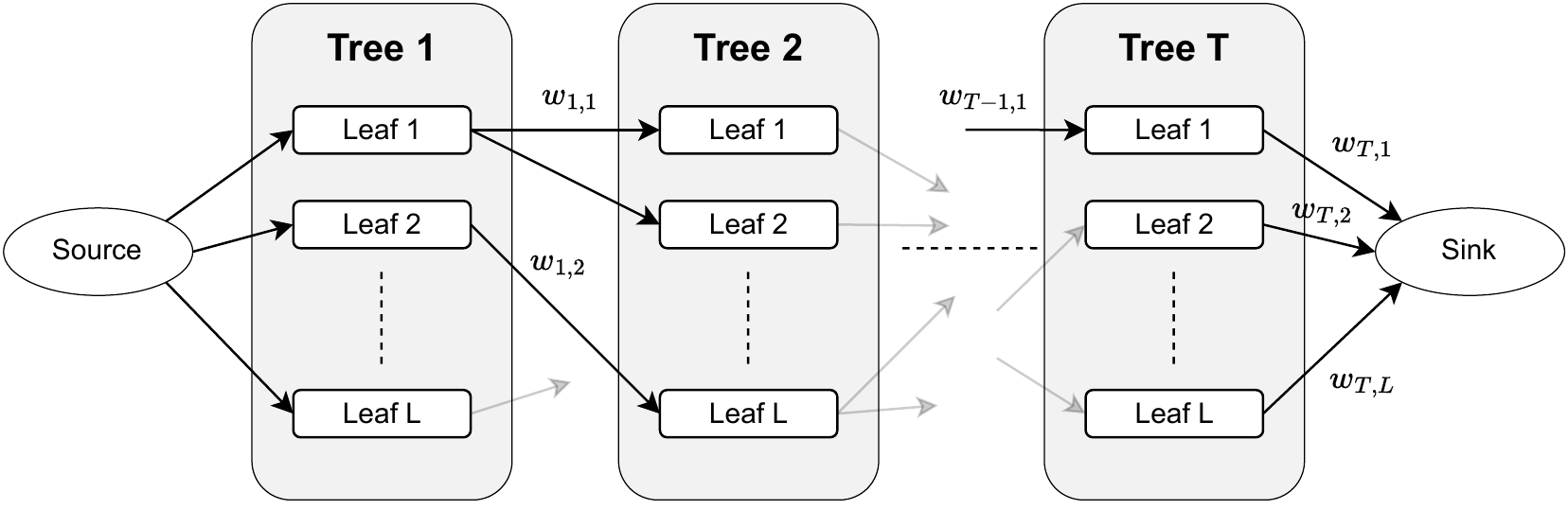}
    \caption{Example of a weighted graph connecting tree leaves in a random forest: two leaves in successive trees are connected only if they have at least one sample in common.}
    \label{fig:weightFlow}
\end{figure}

The CVaR formulation introduced in Section~\ref{sec:cvarFormulation} is based on big-M constraints with constant $W_{\text{max}}$ being an upper bound on the weight of any sample. For nearest-neighbor predictors, the maximum weight of a sample is simply $1/k$. For random forest predictors, the lowest value for $W_{\text{max}}$ is given by:
\begin{equation*}
    \underbar{W} = \max_{x,i} w_i^{RF}(x) = \max_{x,i} \frac{1}{T} \sum_{t=1}^T w_{i, t}^{DT}(x).
\end{equation*}
Calculating this value requires measuring the maximum weights of all samples in all combinations of leaves in the forest, which is intractable even for moderate forests and sample sizes \citep{Vidal2020}. Thus, we look for an upper bound on $\underbar{W}$ that is tighter than the naive bound $W_{\text{max}} = 1$. A first approach is to average over all the maximum tree weights as:
\begin{equation*}
    W_{\text{max}} = \frac{1}{T} \sum_{t=1}^T \max_{x,i} w_{i, t}^{DT}(x) = \frac{1}{T} \sum_{t=1}^T \max_{i} w_{i, t}^{DT}(x_i).
\end{equation*}
However, we noticed in our experiments that this bound is often very close to $1$ and thus does not bring much improvement compared to the naive bound. Hence, we use a second approach based on modeling the weights as a flow over the intersecting leaves of different trees. We create a directed graph where each node $v_{t,l}$ corresponds to the leaf $l$ of tree $t$, as illustrated in Figure~\ref{fig:weightFlow}. For each pair of consecutive trees, we add an arc from node $v_{t,l}$ to $v_{t+1,l'}$ if the two leaf nodes have at least one training sample in common and assign weight $w_{t,l} = 1/n_{t,l}$ to this arc, where $n_{t,l}$ is the number of samples in leaf $l$ of tree~$t$. We also add a sink node connected to all leaf nodes of the last tree with weight $w_{T,l}=1/n_{T,l}$. The bound $W_{\text{max}}$ is finally deduced as the sum of the weights over the arcs of the longest path over the graph.

\section{Supplementary Material: Details of Experimental Setting}
\label{app:exp}
In this section, we introduce in detail the decision problems considered in our numerical experiments. In particular, we present the joint distributions of the contextual information and random parameters and formulate the decision problems as optimization models.
\subsection{Multi-item newsvendor}
In the multi-item newsvendor problem, the decision-maker places orders $\mathbf{z} = [z_1, \dots, z_{d_y}]$ over the product portfolio in the face of uncertain demands. The decision-maker wants to minimize the expected overage and underage costs expressed as:
\begin{equation}
    \label{eq:news}
    \min_{\mathbf{z} \ge 0} \Exp_y \left[ \sum_{j=1}^{d_y} h_j (z_j - y_j)^+ + b_j (y_j - z_j)^+\mid X = x \right],
\end{equation}
where $(z)^+ = \max(z, 0)$. Following \citet{Kallus2022}, the contextual data $X$ is a $2$-dimensional vector drawn from two independent, standard Gaussian distributions. We modify the generation of problem instances to allow different sizes of the product portfolio. The overage and underage costs of product $j$ are set as $h_j = j$ and $b_j = 10 h_j$. We condition the demand of the first half of the product portfolio on the first feature $X_1$ and the demand of the second half on the second feature $X_2$. The demand distribution of product $j$ is thus given by:
\begin{equation*}
    Y_j \mid X = (x_1, x_2) \sim 
    \begin{cases}
        \TruncNorm(3, \exp(x_1)), \text{ if }j \le d_y/2, \\
        \TruncNorm(3, \exp(x_2)), \text{ otherwise.}
    \end{cases}
\end{equation*}
These modifications allow us to generate problem instances of different sizes by varying the number of products $d_y$. Unless stated otherwise, we use $d_y = 20$ in our experiments. Finally, we introduce a budget constraint on the total order quantity over the products:
\begin{equation}
    \label{eq:budget}
    \sum_{j=1}^{d_y} z_j \le K
\end{equation}
with $K = 5 \cdot d_y$. This budget constraint yields a slightly more realistic problem. Further, there is no known analytical solution in this case and the optimal ordering quantity must be computed by solving the linear optimization model given by Equation~\eqref{eq:news} and Constraint~\ref{eq:budget}.

\subsection{Two-stage shipment planning}
The two-stage shipment planning problem is taken directly from \citet[Section EC.6]{Bertsimas2020}. In this problem, an uncertain demand at $d_y=12$ locations is serviced by $d_z=4$ warehouses. The demand depends on $3$-dimensional contextual information, which evolves as an ARMA(2,2) process.

\subsection{(CVaR) shortest path}
We consider a square grid of width $L$ with $L^2$ nodes, and a directed graph connecting each node to its neighbors immediately below and to the right. For any square grid, there are $d_y = 2(l-1)L$ edges with uncertain costs. The decision-maker wants to find the shortest path from the top-left node to the bottom-right node. The decisions $\mathbf{z} \in \{0,1\}^{2(L-1)}$ are binary variables indicating whether arc $j$ is selected or not. We set the grid width to $L=4$ for our experiments with CVaR objectives (same value as \citet{Elmachtoub2020}) and increase the width to $L=8$ for the experiments with expected costs.

The contextual information is a $4$-dimensional vector. In our experiments, we found that the problem setting introduced by \citet{Elmachtoub2020} leads to many symmetric contexts (see Appendix~\ref{app:symSol}). Hence, we modify it in the following way. Each component of the context $x$ is sampled from a uniform distribution $\text{Uniform}(0.5, 1.5)$. The influence of each component of the context vector has an effect on edge $j$ with a probability of $0.5$. The influence matrix $\mathbf{B} \in \{0, 1\}^{d_x, d_y}$ collects the influence of all features on all arcs and is sampled in each experiment. We also randomly generate an additive noise $\varepsilon_{i,j} \sim \text{Uniform}(0,1)$ for each edge $j$ and scenario $i$. In scenario $i$, the cost of edge $j$ is given by:
\begin{equation*}
    y_j^i = \frac{1}{d_x} B_j^\top \cdot \mathbf{x} + \varepsilon_{i,j},
\end{equation*}
where $B_j$ corresponds to the $j$-th column of the influence matrix $\mathbf{B}$. This experimental setting is flexible and permits the use of large-dimensional contextual information while keeping the number of decisions constant, as done in Section~\ref{sec:numRes}.

We consider both expected costs and CVaR objectives with $\alpha = 0.8$. For any node $v \in V$, let $\mathcal{A}_j^-$ and $\mathcal{A}_j^+$ be the set of edges respectively entering and leaving node $j$. The shortest-path problem with expected costs is given by the following optimization model:
\begin{subequations}
    \label{optim:SP}
    \begin{alignat}{2}
        \min_{z}      & \quad && \sum_{i=1}^n w_i (z^\top y_i) \label{optim:SP:obj} \\
        \text{s.t.} &       && \sum_{j \in \mathcal{A}_1^-}z_j = 1 \,, \label{optim:SP:c1} \\
        &       && \sum_{j \in \mathcal{A}_v^-}z_j - \sum_{j \in \mathcal{A}_v^+} z_j = 1 \,, \quad 2 \le v \le \mid V \mid ,\label{optim:SP:c2} \\
        &       && z_j \in \{0, 1\}. \label{optim:SP:c3}
    \end{alignat}
\end{subequations}
Constraint~\eqref{optim:SP:c1} states that at least one edge leaving node $1$ is selected. Constraint~\eqref{optim:SP:c2} ensures the balance of flow at each node. The shortest-path problem with CVaR objective can be obtained by using the convex formulation of \citet{Rockafellar2002}. It is well-known that the solution of the linear relaxation of shortest-path problems coincides with its optimal integer solution. However, this is not the case when the objective is the CVaR.

\section{Supplementary Material: Additional Experiments}
\label{app:numRes}

In this section, we introduce additional experiments to investigate:
\begin{itemize}[topsep=-0.8\parskip, noitemsep]
    \item the occurrence of symmetric contexts in Section~\ref{app:symSol},
    \item the sensitivity of the computational effort and explanations' distance to the hyper-parameters of random forests and nearest-neighbor predictors in Section~\ref{app:hyperparam},
    \item the sensitivity of the explanations to additional parameters in Section~\ref{app:addResults},
    \item the performance of Algorithm~\ref{alg:absCf} compared to the dual formulation presented in Proposition~\ref{prop:absLin} in Section~\ref{app:dual}, and
    \item the impact of spurious features on explanations and their correlation with the alternative contexts in Section~\ref{app:spurious}.
\end{itemize}

\subsection{Symmetric contexts}
\label{app:symSol}
In our experiments, we repeatedly sample new contexts $x_{n+1}$ and $\xAlt$ to obtain decisions $z^*$ and $\zAlt$ and generate diverse explanation problems. We call the two contexts $x_{n+1}$ and $\xAlt$ symmetric when they lead to decisions $z^*$ and $\zAlt$ having the same cost. In that case, $x_{n+1}$ is both a nearest relative and absolute explanation. For each decision problem and set of hyper-parameters, we measure the occurrence of symmetric contexts and present the results in Table~\ref{tab:percSymm}. The table shows a significant difference between the two first problems and the two shortest-path problems. This is expected since shortest-path problems have discrete decisions whereas the newsvendor and shipment planning problems have continuous decisions. It is more likely that two different contexts yield the same decisions and thus the same cost when the decisions are discrete.
\begin{table}[ht!]
    \vspace{-2mm}
    \caption{Occurence of symmetric contexts in the different problems and for varying hyper-parameters [in \%].}
    \label{tab:percSymm}
    \centering
    \begin{sc}
        \resizebox{0.75\textwidth}{!}{\begin{tabular}{*{10}{c}} 
    \toprule   &  & \multicolumn{4}{c}{Relative explanation} & \multicolumn{4}{c}{Absolute explanation} \\ 
    \cmidrule(lr){3-6} \cmidrule(lr){7-10} &  & \multicolumn{2}{c}{RF} & \multicolumn{2}{c}{k-NN} & \multicolumn{2}{c}{RF} & \multicolumn{2}{c}{k-NN} \\ 
    \cmidrule(lr){3-4} \cmidrule(lr){5-6} \cmidrule(lr){7-8} \cmidrule(lr){9-10} &  & $T=100$ & $T=200$ & $k=10$ & $k=30$ & $T=100$ & $T=200$ & $k=10$ & $k=30$ \\ 
    \midrule \multirow{3}{*}{Newsvendor} & $n=50$ & 0 & 0 & 0 & 0 & 0 & 0 & 0 & 0 \\ 
     & $n=100$ & 1 & 0 & 0 & 0 & 1 & 1 & 0 & 0 \\ 
     & $n=200$ & 1 & 1 & 0 & 0 & 2 & 2 & 0 & 0 \\ 
    \cmidrule{2-10} \multirow{3}{*}{Shipment} & $n=50$ & 5 & 2 & 4 & 5 & 5 & 2 & 4 & 5 \\ 
     & $n=100$ & 1 & 1 & 0 & 0 & 1 & 1 & 0 & 0 \\ 
     & $n=200$ & 2 & 1 & 1 & 0 & 2 & 1 & 1 & 0 \\ 
    \cmidrule{2-10} \multirow{3}{*}{Shortest path} & $n=50$ & 47 & 52 & 20 & 51 & 47 & 53 & 21 & 51 \\ 
     & $n=100$ & 43 & 43 & 15 & 40 & 43 & 43 & 15 & 40 \\ 
     & $n=200$ & 42 & 44 & 11 & 35 & 42 & 44 & 11 & 35 \\ 
    \cmidrule{2-10} \multirow{3}{*}{\shortstack{CVaR\\ Shortest path}} & $n=50$ & 68 & 63 & 37 & 70 & 68 & 63 & 37 & 70 \\ 
     & $n=100$ & 71 & 73 & 42 & 70 & 71 & 74 & 42 & 70 \\ 
     & $n=200$ & 61 & 63 & 27 & 52 & 61 & 63 & 27 & 52 \\ 
\bottomrule 
\end{tabular}}
    \end{sc}
    \vspace{-2mm}
\end{table}

The percentage of symmetric contexts depends on the predictor and its hyper-parameters. While the number of trees in the random forest has no significant effect, nearest-neighbor predictors with larger values of $k$ have more symmetric solutions. This is also intuitive since, as $k$ grows toward $n$, the policy converges to a non-contextual decision policy, in which all contexts are symmetric. Finally, note that we exclude all symmetric contexts when reporting computational time throughout the paper.

\subsection{Sensitivity analyses of hyper-parameters}
\label{app:hyperparam}
In this section, we investigate the effect of the hyper-parameters of random forest and nearest-neighbor predictors, namely, the number of trees, the maximum depth of trees, and the number of neighbors.

\paragraph{Computational time} We use the same simulation setting as in Section~\ref{sec:numRes} and vary the size of random forests $T \in \{100, 200\}$ and the number of neighbors $k \in \{10, 30\}$. We show the effect of these hyper-parameters on the computational time in Table~\ref{tab:fullCompTimes}. The time needed to obtain explanations increases for large random forests but remains overall small when the decision problem minimizes expected costs.
\begin{table}[ht]
    \caption{Average computational time for varying predictor hyper-parameters: forest size and $k$ (in s).}
    \label{tab:fullCompTimes}
    \centering
    \begin{sc}
        \resizebox{0.75\textwidth}{!}{\begin{tabular}{*{10}{c}} 
    \toprule   &  & \multicolumn{4}{c}{Relative explanation} & \multicolumn{4}{c}{Absolute explanation} \\ 
    \cmidrule(lr){3-6} \cmidrule(lr){7-10} &  & \multicolumn{2}{c}{RF} & \multicolumn{2}{c}{k-NN} & \multicolumn{2}{c}{RF} & \multicolumn{2}{c}{k-NN} \\ 
    \cmidrule(lr){3-4} \cmidrule(lr){5-6} \cmidrule(lr){7-8} \cmidrule(lr){9-10} &  & $T=100$ & $T=200$ & $k=10$ & $k=30$ & $T=100$ & $T=200$ & $k=10$ & $k=30$ \\ 
    \midrule \multirow{3}{*}{Newsvendor} & $n=50$ & 0.28 & 0.77 & 0.24 & 0.22 & 1.65 & 3.73 & 0.96 & 0.83 \\ 
     & $n=100$ & 0.37 & 1.09 & 1.07 & 1.34 & 3.00 & 6.00 & 5.46 & 6.27 \\ 
     & $n=200$ & 0.48 & 1.44 & 3.71 & 5.01 & 4.41 & 8.12 & 23.90 & 32.78 \\ 
    \cmidrule{2-10} \multirow{3}{*}{Shipment} & $n=50$ & 0.34 & 1.01 & 0.25 & 0.26 & 2.68 & 5.31 & 2.06 & 1.60 \\ 
     & $n=100$ & 0.36 & 1.08 & 1.10 & 2.02 & 5.61 & 9.84 & 9.25 & 13.02 \\ 
     & $n=200$ & 0.47 & 1.52 & 4.12 & 10.91 & 14.35 & 21.52 & 53.32 & 132.17 \\ 
    \cmidrule{2-10} \multirow{3}{*}{Shortest path} & $n=50$ & 0.61 & 2.41 & 0.87 & 0.91 & 1.78 & 5.61 & 2.66 & 2.26 \\ 
     & $n=100$ & 1.02 & 3.99 & 5.34 & 24.09 & 3.67 & 9.75 & 33.77 & 93.18 \\ 
     & $n=200$ & 1.41 & 5.41 & 24.25 & 187.98 & 5.50 & 12.52 & 141.53 & 650.67 \\ 
    \cmidrule{2-10} \multirow{3}{*}{\shortstack{CVaR\\ Shortest path}} & $n=50$ & 6.78 & 26.50 & 1.40 & 1.24 & 7.47 & 31.71 & 4.47 & 2.80 \\ 
     & $n=100$ & 16.27 & 57.78 & 9.45 & 29.82 & 17.88 & 65.75 & 25.60 & 111.65 \\ 
     & $n=200$ & 45.20 & 145.11 & 37.66 & 345.58 & 48.63 & 130.69 & 203.52 & 845.31 \\ 
\bottomrule 
\end{tabular}}
    \end{sc}
\end{table}

We also vary the maximum depth of trees of random forest predictors within $\{3, 4, 5, 6\}$ while keeping the number of trees fixed to $T=100$. We show the impact of the maximum tree depth on the computational time in Table~\ref{tab:depthCompTimes}. While increasing the maximum tree depth increases the computational time, explanations can still be obtained quickly within a few seconds to a few minutes on the considered problems.
\begin{table}[ht]
    \caption{Average computational time for varying maximum tree depth [in s].}
    \label{tab:depthCompTimes}
    \centering
    \begin{sc}
        \resizebox{0.75\textwidth}{!}{\begin{tabular}{*{10}{c}} 
    \toprule   &  & \multicolumn{4}{c}{Relative explanation} & \multicolumn{4}{c}{Absolute explanation} \\ 
    \cmidrule(lr){3-6} \cmidrule(lr){7-10} \multicolumn{2}{r}{Max. tree depth} & $3$ & $4$ & $5$ & $6$ & $3$ & $4$ & $5$ & $6$ \\ 
    \midrule \multirow{3}{*}{Newsvendor} & $n=50$ & 0.13 & 0.29 & 0.51 & 0.86 & 0.94 & 1.72 & 2.79 & 4.16 \\ 
     & $n=100$ & 0.16 & 0.34 & 0.66 & 1.19 & 1.71 & 2.73 & 4.07 & 6.84 \\ 
     & $n=200$ & 0.21 & 0.42 & 0.86 & 1.65 & 2.49 & 4.09 & 6.48 & 10.58 \\ 
    \cmidrule{2-10} \multirow{3}{*}{Shipment} & $n=50$ & 0.15 & 0.32 & 0.60 & 1.06 & 1.62 & 2.57 & 4.50 & 6.81 \\ 
     & $n=100$ & 0.17 & 0.40 & 0.69 & 1.41 & 3.60 & 6.24 & 8.69 & 14.08 \\ 
     & $n=200$ & 0.20 & 0.46 & 0.98 & 2.42 & 10.04 & 14.94 & 21.97 & 33.26 \\ 
    \cmidrule{2-10} \multirow{3}{*}{Shortest path} & $n=50$ & 0.24 & 0.78 & 1.66 & 2.50 & 1.08 & 2.22 & 3.85 & 6.00 \\ 
     & $n=100$ & 0.33 & 1.10 & 3.01 & 6.01 & 1.76 & 3.37 & 7.28 & 12.59 \\ 
     & $n=200$ & 0.36 & 1.45 & 3.96 & 11.71 & 3.34 & 5.78 & 10.69 & 25.40 \\ 
    \cmidrule{2-10} \multirow{3}{*}{\shortstack{CVaR\\ Shortest path}} & $n=50$ & 2.54 & 4.78 & 8.66 & 14.09 & 3.25 & 6.01 & 10.07 & 17.07 \\ 
     & $n=100$ & 6.41 & 18.11 & 37.74 & 61.68 & 6.15 & 20.57 & 44.61 & 73.85 \\ 
     & $n=200$ & 16.32 & 40.97 & 97.54 & 242.87 & 13.79 & 40.76 & 121.58 & 220.07 \\ 
\bottomrule 
\end{tabular}}
    \end{sc}
\end{table}

\paragraph{Distance of explanations}
We now measure the distance between the obtained explanations and the initial context $x_{n+1}$ using the $l_1$ norm. We show how the average distance varies with the model hyperparameters in Table~\ref{tab:fullDist}. Interestingly, the hyper-parameters of the predictors have little effect on the distance of explanations.
\begin{table}[ht]
    \caption{Average distance of explanations for varying predictor hyper-parameters: forest size and $k$ [in s].}
    \label{tab:fullDist}
    \centering
    \begin{sc}
        \resizebox{0.75\textwidth}{!}{\begin{tabular}{*{10}{c}} 
    \toprule   &  & \multicolumn{4}{c}{Relative explanation} & \multicolumn{4}{c}{Absolute explanation} \\ 
    \cmidrule(lr){3-6} \cmidrule(lr){7-10} &  & \multicolumn{2}{c}{RF} & \multicolumn{2}{c}{k-NN} & \multicolumn{2}{c}{RF} & \multicolumn{2}{c}{k-NN} \\ 
    \cmidrule(lr){3-4} \cmidrule(lr){5-6} \cmidrule(lr){7-8} \cmidrule(lr){9-10} &  & $T=100$ & $T=200$ & $k=10$ & $k=30$ & $T=100$ & $T=200$ & $k=10$ & $k=30$ \\ 
    \midrule \multirow{3}{*}{Newsvendor} & $n=50$ & 0.3 & 0.28 & 0.39 & 0.36 & 0.88 & 0.88 & 0.88 & 0.83 \\ 
     & $n=100$ & 0.37 & 0.37 & 0.44 & 0.44 & 1.09 & 1.07 & 1.12 & 1.13 \\ 
     & $n=200$ & 0.27 & 0.27 & 0.37 & 0.37 & 0.82 & 0.85 & 0.93 & 0.96 \\ 
    \cmidrule{2-10} \multirow{3}{*}{Shipment} & $n=50$ & 0.28 & 0.26 & 0.38 & 0.39 & 0.67 & 0.6 & 0.72 & 0.65 \\ 
     & $n=100$ & 0.21 & 0.19 & 0.34 & 0.46 & 0.66 & 0.7 & 0.86 & 0.9 \\ 
     & $n=200$ & 0.19 & 0.2 & 0.47 & 0.47 & 0.59 & 0.65 & 0.94 & 1.01 \\ 
    \cmidrule{2-10} \multirow{3}{*}{Shortest path} & $n=50$ & 0.03 & 0.04 & 0.04 & 0.04 & 0.04 & 0.05 & 0.06 & 0.04 \\ 
     & $n=100$ & 0.05 & 0.05 & 0.04 & 0.04 & 0.06 & 0.07 & 0.06 & 0.05 \\ 
     & $n=200$ & 0.06 & 0.06 & 0.04 & 0.03 & 0.07 & 0.07 & 0.06 & 0.04 \\ 
    \cmidrule{2-10} \multirow{3}{*}{\shortstack{CVaR\\ Shortest path}} & $n=50$ & 0.05 & 0.06 & 0.05 & 0.03 & 0.06 & 0.07 & 0.06 & 0.04 \\ 
     & $n=100$ & 0.04 & 0.04 & 0.04 & 0.04 & 0.04 & 0.04 & 0.05 & 0.04 \\ 
     & $n=200$ & 0.04 & 0.04 & 0.03 & 0.04 & 0.05 & 0.05 & 0.04 & 0.04 \\ 
\bottomrule 
\end{tabular}}
    \end{sc}
\end{table}

\subsection{Additional results on synthetic data}
\label{app:addResults}

\paragraph{Sparsity of explanations.} In Section~\ref{sec:numRes}, we investigate shortest-path problems with a large number of features and training samples and observed that explanations can be obtained in a short time. The average number of features changed in explanations obtained for $n=500$ training samples is shown in Table~\ref{tab:sparsity}. The table shows that the explanations obtained are sparse: only a few features are changed, even when the context has a large number of dimensions. This can be explained by our use of the $\ell_1$ norm to measure the distance of explanations to the original contexts, which is known to favor sparse explanations.
\begin{table}[ht]
    \caption{Average number of features changed to obtain explanations.}
    \label{tab:sparsity}
    \centering
    \begin{sc}
        \resizebox{0.6\textwidth}{!}{\begin{tabular}{*{3}{c}} 
    \toprule
     Number of features $d_x$ & \multicolumn{2}{c}{Number of features changed (percentage)}\\
     \cmidrule(lr){2-3}  & Relative explanations  & Absolute explanations \\
     \midrule
    5    & 2.8 (43\%)          & 2.5 (49\%)   \\
    10   & 3.2 (32\%)          & 3.6 (36\%)   \\
    25   & 2.3 (9\%)           & 2.5 (10\%)   \\
    50   & 1.6 (3.3\%)         & 1.9 (3.8\%)  \\
    100  & 2.0 (2.0\%)         & 2.3 (2.3\%)  \\
    500  & 3.6 (0.7\%)         & 3.6 (0.7\%)  \\
    \bottomrule
\end{tabular}}
    \end{sc}
\end{table}

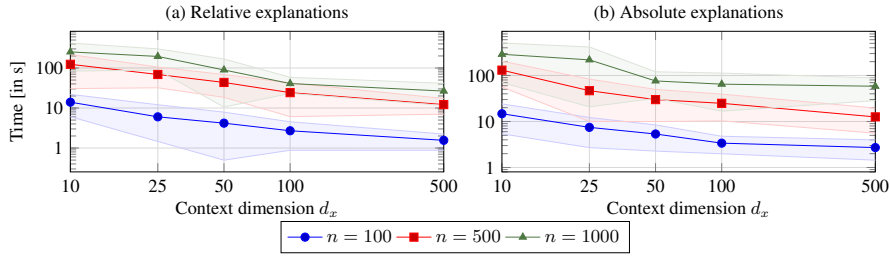
\begin{figure}[ht!]
    \centering
    \resizebox{0.7\linewidth}{!}{\tikzset{dgStyle/.style={darkgreen, mark=triangle*,mark options={fill=darkgreen}}}

\begin{tikzpicture}
	\begin{groupplot}[
		group style={
			group name=my plots,
			group size=2 by 1,
			xlabels at=edge bottom,
			ylabels at=edge left
		},
		height = 4cm,
		width  = 8cm,
        ymode=log,
        xmin=10, xmax=500,
        xmode=log,
        xtick = {10, 25, 50, 100, 500},
        xticklabels = {10, 25, 50, 100, 500},
        ytick = {0.1, 1, 10, 100, 1000},
        yticklabels = {0.1, 1, 10, 100, 1000},
		xlabel = {Context dimension $d_x$},
		ylabel = {Time [in s]},
        ylabel shift = -2 pt,
        xlabel shift = -2 pt
		]
		\nextgroupplot[title = {(a) Relative explanations}, title style={yshift=-5pt},
                        font = \small]
        %%%%%%%%%%% n=100 %%%%%%%%%%%
		\addplot+[blue] table [x index = {0}, y index = {1}, col sep=comma]{plots/cvar_path_feat_sens.csv};
        \addplot+[name path=C, blue!20, mark=none, forget plot] table [x index = {0}, y index = {2}, col sep=comma]{plots/cvar_path_feat_sens.csv};
        \addplot+[name path=D, blue!20, mark=none, forget plot] table [x index = {0}, y index = {3}, col sep=comma]{plots/cvar_path_feat_sens.csv};
        \addplot[blue, fill opacity=0.05, forget plot] fill between[of=C and D];
        %%%%%%%%%%% n=500 %%%%%%%%%%%
		\addplot+[red] table [x index = {0}, y index = {7}, col sep=comma]{plots/cvar_path_feat_sens.csv};
        \addplot+[name path=C, red!20, mark=none, forget plot] table [x index = {0}, y index = {8}, col sep=comma]{plots/cvar_path_feat_sens.csv};
        \addplot+[name path=D, red!20, mark=none, forget plot] table [x index = {0}, y index = {9}, col sep=comma]{plots/cvar_path_feat_sens.csv};
        \addplot[red, fill opacity=0.05, forget plot] fill between[of=C and D];
        %%%%%%%%%%% n=1000 %%%%%%%%%%%
		\addplot+[dgStyle] table [x index = {0}, y index = {13}, col sep=comma]{plots/cvar_path_feat_sens.csv};
        \addplot+[name path=C, darkgreen!20, mark=none, forget plot] table [x index = {0}, y index = {14}, col sep=comma]{plots/cvar_path_feat_sens.csv};
        \addplot+[name path=D, darkgreen!20, mark=none, forget plot] table [x index = {0}, y index = {15}, col sep=comma]{plots/cvar_path_feat_sens.csv};
        \addplot[darkgreen, fill opacity=0.05, forget plot] fill between[of=C and D];
		
		\nextgroupplot[title = {(b) Absolute explanations}, title style={yshift=-5pt}, font = \small,
			legend style={at={(-0.125,-0.35)},anchor=north},
			legend columns=4]
        %%%%%%%%%%% n=100 %%%%%%%%%%%
		\addplot+[blue] table [x index = {0}, y index = {4}, col sep=comma]{plots/cvar_path_feat_sens.csv};
        \addlegendentry{$n=100$}
        \addplot+[name path=C, blue!20, mark=none, forget plot] table [x index = {0}, y index = {5}, col sep=comma]{plots/cvar_path_feat_sens.csv};
        \addplot+[name path=D, blue!20, mark=none, forget plot] table [x index = {0}, y index = {6}, col sep=comma]{plots/cvar_path_feat_sens.csv};
        \addplot[blue, fill opacity=0.05, forget plot] fill between[of=C and D];
        %%%%%%%%%%% n=500 %%%%%%%%%%%
		\addplot+[red] table [x index = {0}, y index = {10}, col sep=comma]{plots/cvar_path_feat_sens.csv};
        \addlegendentry{$n=500$}
        \addplot+[name path=C, red!20, mark=none, forget plot] table [x index = {0}, y index = {11}, col sep=comma]{plots/cvar_path_feat_sens.csv};
        \addplot+[name path=D, red!20, mark=none, forget plot] table [x index = {0}, y index = {12}, col sep=comma]{plots/cvar_path_feat_sens.csv};
        \addplot[red, fill opacity=0.05, forget plot] fill between[of=C and D];
        %%%%%%%%%%% n=1000 %%%%%%%%%%%
		\addplot+[dgStyle] table [x index = {0}, y index = {16}, col sep=comma]{plots/cvar_path_feat_sens.csv};
        \addlegendentry{$n=1000$}
        \addplot+[name path=C, darkgreen!20, mark=none, forget plot] table [x index = {0}, y index = {17}, col sep=comma]{plots/cvar_path_feat_sens.csv};
        \addplot+[name path=D, darkgreen!20, mark=none, forget plot] table [x index = {0}, y index = {18}, col sep=comma]{plots/cvar_path_feat_sens.csv};
        \addplot[darkgreen, fill opacity=0.05, forget plot] fill between[of=C and D];

	\end{groupplot}
\end{tikzpicture}}
    \caption{c-SP: computational time on large instances (standard deviation is shown in shaded area).}
    \label{fig:cvarFeatSens}
\end{figure}

\paragraph{CVaR shortest path.} We present the sensitivity of the computational time to obtain relative and absolute explanations of the CVaR shortest-path problem with an increasing number of features in Figure~\ref{fig:cvarFeatSens}. The results are similar to those on the shortest path with expected costs: our methods provide explanations in a reasonable time even when the contextual information consists of many informative features. In fact, as in Figure~\ref{fig:featSens}, the computational time tends to decrease as the number of features increases.

\subsection{Absolute explanations: Iterative algorithm vs. dual reformulation}
\label{app:dual}
\begin{figure}[ht!]
	\centering
	\includegraphics[trim=2mm 2mm 2mm 2mm, clip, width=0.8\linewidth]{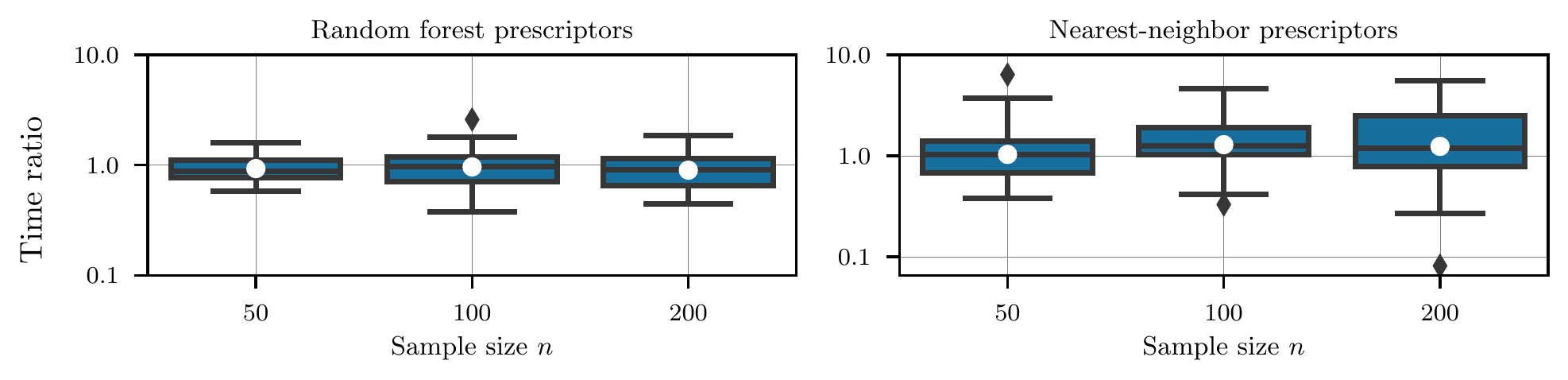}
	\caption{Relative solving time for obtaining absolute explanations: iterative approach relative to dual approach}
	\label{fig:dualVsIter}
    \vspace{-.3cm}
\end{figure}

Throughout the paper, we use the iterative procedure given in Algorithm~\ref{alg:absCf} implemented efficiently with callbacks and lazy constraints to obtain absolute explanations. In this section, we compare the performance of this algorithm compared to the dual reformulation given in Proposition~\ref{prop:absLin}. We focus on the shortest-path problem with expected costs since it is the linear decision problem with the largest computational time for absolute explanations. We show the solving time of Algorithm~\ref{alg:absCf} relative to the solving time of the dual reformulation in Figure~\ref{fig:dualVsIter}.

Algorithm~\ref{alg:absCf} provides the best performance for random forests with large training sets, whereas the dual reformulation is the fastest on average for nearest-neighbor predictors. However, it is interesting to observe that the distribution of the time ratios has a large variance, especially for nearest-neighbor predictors. We leave it to future work to characterize the situations in which Algorithm~\ref{alg:absCf} or the dual reformulation will provide explanations in the shortest time.

\subsection{Spurious features and explanations}
\label{app:spurious}
In this section, we extend our analyses regarding the correlation of explanations by including spurious features. We focus on the newsvendor problem and introduce two additional features that are completely independent of the demand. The spurious features are also set to follow standard normal distributions. We repeatedly generate contexts and determine explanations and measure two indicators: (1)~ the correlation between explanations and alternative contexts as in Section~\ref{sec:numRes}, and (2)~the percentage of changes made on relevant features rather than spurious ones, calculated as $p = \frac{\lVert \xAlt_{(1:2)} - x^{n+1}_{(1:2)} \lVert_2}{\lVert \xAlt - x_{n+1} \rVert_2}$.

\begin{figure}[ht]
	\centering
	\includegraphics[trim=2mm 2mm 2mm 2mm, clip, width=0.8\linewidth]{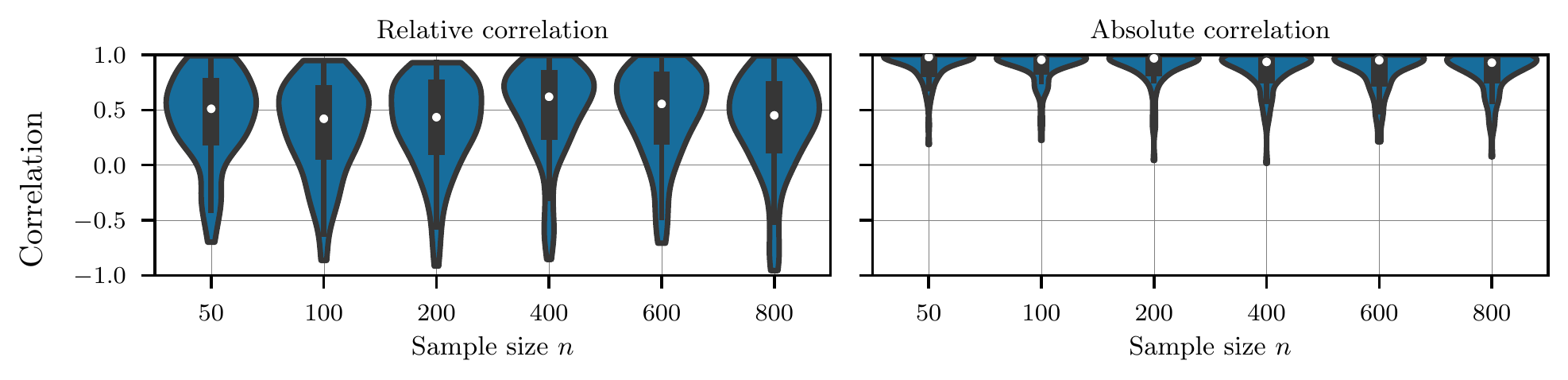}
	\caption{Correlation between obtained explanations and alternative contexts with spurious features.}
	\label{fig:appCorr}
    \vspace{-.3cm}
\end{figure}

\begin{figure}[ht]
	\centering
	\includegraphics[trim=2mm 2mm 2mm 2mm, clip, width=0.8\linewidth]{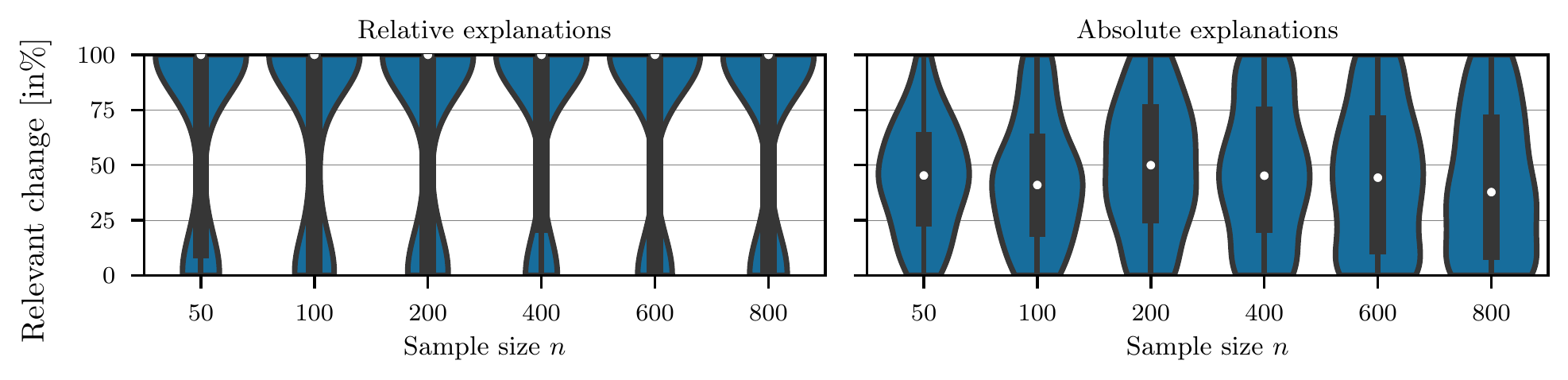}
	\caption{Percentage of change made on relevant rather than spurious features.}
	\label{fig:appPerc}
    \vspace{-.3cm}
\end{figure}

We focus on random forest predictors and vary the sample size $n \in \{50, 100, 200, 400, 600, 800 \}$. The correlation between explanations and the alternative contexts is shown in Figure~\ref{fig:appCorr}, and the percentage of changes made on relevant features is shown in Figure~\ref{fig:appPerc}. The figures suggest that relative explanations tend to modify only relevant features and avoid spurious ones. However, they do not always move the explanation toward the alternative context. On the other hand, absolute explanations tend to modify all features equally, but stay almost completely aligned with the direction of the alternative contexts. These observations remain valid for all sample sizes investigated.

\end{document}